%% file: main.tex
\documentclass[runningheads]{llncs}
\usepackage[T1]{fontenc}
\usepackage{graphicx}
\usepackage{booktabs} 
\usepackage{amsmath} 
\usepackage{comment}

\usepackage{array}
\usepackage[table]{xcolor}
\usepackage{tabularx}
\usepackage{tikz}
\usetikzlibrary{shapes.geometric}
\usepackage{amsfonts}
\usepackage{hyperref} 
\usepackage{float}
\usepackage{subcaption}
\usepackage{csquotes}

\begin{document}
\title{From Rules to Rewards: Reinforcement Learning for Interest Rate Adjustment in DeFi Lending}
%Short Paper: Infinite Scalability for Ethereum: Evaluating ZK Rollups in Real-Time Applications
%
\titlerunning{Reinforcement Learning in DeFi Lending}
\author{Hanxiao Qu\inst{1} \and
Krzysztof M. Gogol\inst{1}\thanks{Corresponding author: \email{gogol@ifi.uzh.ch}} \and
Florian Grötschla\inst{2} \and \newline
Claudio J. Tessone\inst{1,3}}
\institute{University of Zurich \and ETH Zurich \and UZH Blockchain Center}
\authorrunning{Hanxiao Qu, Krzysztof Gogol, Florian Grötschla, Claudio Tessone}
\maketitle              % typeset the header of the contribution
\begin{abstract}
\input{sections/00Abstract}

\end{abstract}
\input{sections/01Introduction}
\input{sections/02Background}
\input{sections/03Data}

\input{sections/04Evaluation}

\input{sections/08Discussion}

\input{sections/09Conclusions}

 \bibliographystyle{splncs04}
 \bibliography{main}

\appendix
\input{sections/97RelatedWork}
\input{sections/98DataPreProcessing}
\input{sections/99Appendix}

\end{document}

%% file: sections/00Abstract.tex
Decentralized Finance (DeFi) lending enables permissionless borrowing via smart contracts. However, it faces challenges in optimizing interest rates, mitigating bad debt, and improving capital efficiency. Rule-based interest-rate models struggle to adapt to dynamic market conditions, leading to inefficiencies.
This work applies Offline Reinforcement Learning (RL) to optimize interest rate adjustments in DeFi lending protocols. Using historical data from Aave protocol, we evaluate three RL approaches: Conservative Q-Learning (CQL), Behavior Cloning (BC), and TD3 with Behavior Cloning (TD3-BC).
TD3-BC demonstrates superior performance in balancing utilization, capital stability, and risk, outperforming existing models. It adapts effectively to historical stress events like the May 2021 crash and the March 2023 USDC depeg, showcasing potential for automated, real-time governance.
%
%Decentralized Finance (DeFi) has revolutionized financial markets by enabling permissionless lending and borrowing through smart contracts. However, DeFi lending protocols face critical challenges, including optimizing interest rate mechanisms, minimizing bad debt risks, and ensuring capital efficiency. Traditional rule-based interest rate models often fail to adapt dynamically to evolving market conditions, leading to inefficiencies in liquidity provision and borrowing costs.
%
%This thesis explores the application of Offline Reinforcement Learning (RL) to optimize interest rate adjustments in DeFi lending protocols. Using historical data from Aave V2 and Aave V3, we develop and evaluate an RL-based optimization framework trained on real-world lending behavior. The study compares Conservative Q-Learning (CQL), Behavior Cloning (BC), and Twin Delayed Deep Deterministic Policy Gradient with Behavior Cloning (TD3-BC) in order to identify the most effective model for optimizing lending rates.
%
%Our results show that TD3-BC outperforms the other models in balancing utilization efficiency, capital stability, and risk mitigation, offering a viable alternative to traditional DeFi lending rate models. The RL framework effectively adapts to historical stress events, such as the May 2021 market crash and the March 2023 USDC depeg, demonstrating its potential for real-time governance automation.

%% file: sections/01Introduction.tex
\section{Introduction}
\label{sec:introduction}

%Decentralized Finance (DeFi) has emerged as one of the most transformative applications of blockchain technology, offering financial services such as lending, borrowing, and yield farming without reliance on traditional intermediaries. By leveraging smart contracts, DeFi platforms eliminate the need for banks and centralized institutions, enabling fully transparent, permissionless, and programmable financial interactions. This innovation has significantly reshaped financial markets, providing open access to financial services for anyone with an internet connection. Unlike traditional finance, where lending and borrowing are regulated by financial institutions and subject to stringent approval processes, DeFi enables direct, automated transactions between users. The decentralized nature of DeFi allows for an inclusive, censorship-resistant financial system, removing barriers to entry while increasing transparency in the management of financial assets.

%Since its inception, DeFi has experienced exponential growth, with Total Value Locked (TVL)-a key metric representing the capital deposited in DeFi protocols-expanding from less than one billion dollars in 2019 to over two hundred and fifty billion dollars at its peak in late 2021. This growth was driven by several factors, including increased adoption of Ethereum-based smart contracts, the rise of algorithmic stablecoins, and the expansion of liquidity incentives through yield farming. 

Decentralized Finance (DeFi) lending has become a core component of the DeFi ecosystem, allowing two groups of users to benefit: liquidity providers (LPs), who deposit idle assets to earn interest, and borrowers, who obtain liquidity without the need for credit approval~\cite{gogol2024sokdefi,schaer2023defimarkets,werner2022sok}. 
%However, as DeFi lending markets have grown, so too have their challenges. While early DeFi lending protocols relied on simple interest rate models and static governance parameters, the rapid expansion of decentralized lending has highlighted inefficiencies in capital allocation, liquidity management, and risk mitigation~\cite{gudgeon2020defiprotocolsloanablefunds}.
A defining feature of DeFi lending is that the logic governing interest rates is implemented directly in smart contracts. Unlike in traditional finance, where central banks and financial institutions set borrowing costs based on macroeconomic conditions, DeFi lending rates fluctuate dynamically based on liquidity supply and demand conditions~\cite{gudgeon2020defiprotocolsloanablefunds}. %This approach ensures that capital is efficiently allocated within the system, but it also introduces volatility and unpredictability in borrowing costs. During periods of high utilization, interest rates can spike dramatically, making borrowing prohibitively expensive. Conversely, when utilization is low, liquidity remains idle, leading to inefficient capital allocation and reduced incentives for depositors. The inability of current DeFi lending protocols to dynamically optimize interest rates in response to real-time market conditions has led to inefficiencies that impact both lenders and borrowers.

Lending protocols account for 30-40\% of Total Value Locked (TVL) in DeFi. In early 2021, DeFi lending markets saw a dramatic increase in capital inflows, fueled by the growing adoption of decentralized exchanges, stablecoins, and cross-chain liquidity solutions. By November 2021, DeFi TVL reached an all-time high, with Aave alone surpassing 25b\$ in locked assets, making it the largest lending platform at the time~\cite{2024DeFiLlama}. However, this growth was not without setbacks. Major market disruptions, such as the Terra-LUNA collapse in May 2022 and the FTX insolvency in November 2022, triggered mass liquidity exits, causing sharp TVL drop in early 2023. These market events exposed vulnerabilities in existing lending models, particularly during extreme volatility. %During these downturns, interest rate models failed to adapt quickly enough to shifting market conditions, leading to liquidity shortages, higher borrowing costs, and an increase in bad debt accumulation.

%One of the most pressing issues in DeFi lending is the challenge of balancing liquidity efficiency with risk management. 
Lending protocols rely on overcollateralization to ensure the security of loans, meaning that borrowers must deposit assets worth more than the amount they wish to borrow. While this mechanism reduces default risk, it also creates inefficiencies, as large amounts of capital remain locked in smart contracts rather than being actively utilized~\cite{gogol2024sokdefi}. Moreover, extreme market fluctuations can lead to liquidations that fail to recover outstanding debt, exposing lending protocols to systemic risk. The USDC depegging event in March 2023 and the cascading contagion from the FTX collapse in late 2022 highlighted the fragility of current risk management strategies. In both cases, borrowers faced unexpected liquidation events, leading to a wave of forced sell-offs that further exacerbated market instability. Protocol governance mechanisms, which often rely on community voting to adjust lending parameters, proved too slow to respond to these rapid changes, compounding liquidity crises and increasing systemic exposure.

The limitations of existing DeFi lending models have led to a growing interest in the use of machine learning techniques to optimize interest rate mechanisms, improve liquidity allocation, and improve risk management strategies. Unlike traditional rule-based approaches, machine learning models can analyze historical market data, identify patterns, and dynamically adjust lending parameters in response to real-time changes in market conditions. Among machine learning techniques, reinforcement learning is particularly well-suited for optimizing DeFi lending because it allows an agent to learn from past market conditions and take actions that maximize long-term capital efficiency while minimizing risk exposure. %By continuously adapting to new data, reinforcement learning can provide a more responsive and effective approach to managing interest rates, balancing liquidity supply and demand, and mitigating bad debt risks.

%Reinforcement learning has already been applied in various financial markets, including portfolio optimization, algorithmic trading, and risk modeling. However, its application in DeFi lending remains relatively unexplored. Recent advancements in offline reinforcement learning, which allows models to be trained on historical data without requiring direct real-time interaction with the lending protocol, have made AI-driven interest rate optimization more viable. Unlike traditional models that rely on static utilization-based formulas, reinforcement learning-based frameworks can dynamically adjust interest rates based on actual borrowing demand, improving liquidity management and borrower affordability. These models also contribute to better risk management by predicting potential liquidation events and optimizing collateralization requirements accordingly.

\subsection*{Contributions}
 This work presents how different Reinforcement Learning (RL) approaches-Conservative Q-Learning (CQL), Behavior Cloning (BC), and Twin Delayed Deep Deterministic Policy Gradient with Behavior Cloning (TD3-BC)-can optimize DeFi lending strategies. 
This RL-driven framework learns from historical data (Aave v1, v2) and generates policies that improve capital efficiency in a decentralized setting. %The code and data are publicly available at the Author's GitHub repositories.
 %The research aims to assess the effectiveness of these models in balancing capital efficiency, borrower costs, and risk mitigation, ultimately contributing to the development of more robust and adaptive DeFi lending protocols. 
 The contributions are:

\begin{enumerate}

\item Developing an RL-based optimization framework that can learn from historical market data to automatically adjust interest rates in response to dynamic utilization changes, capital constraints, and borrower demand.
\item Comparing the efficacy of three RL approaches-CQL, BC, and TD3-BC-in enhancing capital efficiency, liquidity utilization, and risk mitigation across different lending market conditions.
\item Evaluating the ability of RL models to respond to extreme market events, such as the May 2021 market crash, the March 2023 USDC depeg, and the November 2022 FTX contagion, by examining how the learned policies react to stress-test scenarios.
\end{enumerate}

%Unlike traditional rule-based lending rate models, this RL-driven framework learns from historical data and generates policies that improve capital efficiency in a decentralized setting. The code and data are publicly available at the Author's GitHub repositories.

\begin{comment}

\subsection*{Paper Structure}

This works structured as follows:
\begin{itemize}
\item \textbf{Chapter 2 (Theoretical Foundations)} introduces the fundamental principles of DeFi lending markets, interest rate modeling, risk management strategies, and reinforcement learning methodologies.
\item \textbf{Chapter 3 (Methodology)} outlines the data collection process from AaveScan, feature engineering, reward function design, and model training pipeline for RL-based lending rate optimization.
\item \textbf{Chapter 4 (Empirical Analysis)} presents an in-depth comparative evaluation of the trained models, analyzing their performance across liquidity metrics, interest rate dynamics, and stress-test robustness.
\item \textbf{Chapter 5 (Final Considerations)} synthesizes our findings, presents key conclusions, and discusses future research directions for RL-driven DeFi automation.

\end{itemize}

\end{comment}

%% file: sections/02Background.tex
\section{Background}
    \label{sec:background}

\subsection{DeFi Lending}
\subsubsection{Interest Rate Mechanisms:}

Users of DeFi lending protocols can deposit their assets to earn interest, or borrow assets by providing collateral. The process is facilitated by a smart contract that determines interest rates algorithmically based on the supply and demand of the assets.
   % \item Collateral: Assets pledged by borrowers to secure the loan, which can be liquidated if the loan's value exceeds a certain threshold.
  %  \item Liquidation Mechanisms: Processes that automatically sell the borrower's collateral if its value falls below the required collateralization ratio, ensuring the protocol's solvency.
 A common model employed is the utilization rate model, in which the interest rate is a function of the proportion of lent assets relative to the total available assets.
 
\noindent
\textbf{Utilization Rate (U):}
\begin{equation}
    U = \frac{\text{Total Value Borrowed}}{\text{Total Value Supplied}}
\end{equation}

%\textit{Note: The utilization rate reflects the demand for a particular asset within the protocol.}
\noindent
\textbf{Borrowing Interest Rate ($R_b$):}
\begin{equation}
    R_b =
    \begin{cases} 
      r_{\text{base}} + U \times \text{slope}_1, & \text{if } U \leq U^* \\
      r_{\text{base}} + U^* \times \text{slope}_1 + (U - U^*) \times \text{slope}_2, & \text{if } U > U^*
    \end{cases}
\end{equation}

where:
\begin{itemize}
    \item $r_{\text{base}}$ is the base interest rate.
    \item $U^*$ is the optimal utilization rate (kink point).
    \item $\text{slope}_1$ and $\text{slope}_2$ are the interest rate slopes before and after the kink point, respectively.
\end{itemize}

%\textit{Note: This piecewise function ensures that interest rates increase more steeply once the utilization rate surpasses the optimal point, incentivizing more liquidity provision.}

\noindent
\textbf{Deposit Interest Rate ($R_d$):}
\begin{equation}
    R_d = R_b \times U \times (1 - \text{Reserve Factor})
\end{equation}
where Reserve Factor is a percentage of the interest paid by borrowers. It is accumulated by the protocol for reserves or insurance purposes.

%\textit{Note: The deposit interest rate is derived from the borrowing rate, adjusted by the utilization rate and reserve factor, ensuring that lenders are compensated proportionally to the demand for their assets.}

%\textbf{Case: Aave Protocol}
%Aave, a prominent DeFi lending platform, utilizes a dynamic interest rate model with a kink point to balance supply and demand. For instance, if the optimal utilization rate ($U^*$) is set at 80\%, the interest rate increases moderately until this point. Beyond 80\% utilization, the interest rate slope becomes steeper, significantly increasing borrowing costs to attract more liquidity into the pool.
%\textit{Note: Aave's adaptive interest rate model helps maintain liquidity and stability within the protocol.}

\subsubsection{Risk Mitigation:}
Due to the lack of credit-scoring of borrowers, the lending protocols apply \textbf{overcollateralization}, requiring borrowers to lock collateral whose value exceeds the borrowed amount. 

Liquidation is the process in which a borrower's collateral is automatically sold to repay the outstanding loan, thereby preserving the financial stability of the protocol. 
A \textbf{liquidation threshold} is a predefined collateralization ratio set by the protocol, below which a borrower's position becomes eligible for liquidation.
The liquidation process may also involve a penalty fee, which incentivizes third-party liquidators and discourages risky borrowing behavior. It is often automated with keeper networks or liquidation bots, ensure rapid response to price fluctuations.

Protocols often complement these safeguards with \textbf{stability fees}---charges levied on borrowers to compensate for the risk of holding volatile collateral. Insurance funds and protocol reserves provide a secondary layer of protection by covering potential shortfalls during extreme market events.

\subsection{Reinforcement Learning in Financial Applications }

Reinforcement Learning (RL) is a subfield of machine learning where an agent interacts with an environment, receiving rewards for actions that maximize long-term benefits. %In DeFi lending, RL can be applied to:
%\begin{itemize}
%    \item Dynamically adjust lending rates based on market conditions.
%    \item Optimize liquidation thresholds to minimize bad debt risks.
%    \item Identify optimal utilization levels to balance supply and demand.
%\end{itemize}
%\textbf{Workflow of RL in Finance}
The RL workflow in finance follows a structured learning process. The agent explores the environment, takes actions, and receives rewards based on performance. It continuously refines its strategy to optimize financial outcomes. RL is typically framed as a Markov Decision Process (MDP), defined as a tuple:

\begin{equation}
    \mathcal{M} = (\mathcal{S}, \mathcal{A}, P, R, \gamma)
\end{equation}
where:
\begin{itemize}
    \item $\mathcal{S}$ is the state space, representing market conditions such as interest rates, liquidity, and utilization rates.
    \item $\mathcal{A}$ is the action space, defining parameter adjustments (e.g., modifying interest rates or collateral requirements).
    \item $P(s' | s, a)$ is the transition probability function, describing the stochastic evolution of market states.
    \item $R(s, a)$ is the reward function, which quantifies the objective (e.g., maximizing liquidity while minimizing bad debt).
    \item $\gamma \in (0,1]$ is the discount factor, weighting future rewards.
\end{itemize}
The agent's objective is to learn a policy $\pi(a | s)$ that maximizes the expected cumulative reward:

\begin{equation}
    J(\pi) = \mathbb{E} \left[ \sum_{t=0}^{\infty} \gamma^t R(s_t, a_t) \right]
\end{equation}
\noindent
where the expectation is taken over the state-action trajectory induced by $\pi$.
RL models used in finance rely on different training methodologies and policy-learning mechanisms. The mathematical descriptions of key models follow.

\vspace{0.5pt}
\noindent
\textbf{Q-Learning:} Q-Learning is a model-free RL algorithm where the agent learns the optimal action-selection policy using the Bellman equation \cite{sutton2018reinforcement}:

\begin{equation}
    Q(s, a) \leftarrow Q(s, a) + \alpha \left[ r + \gamma \max_{a'} Q(s', a') - Q(s, a) \right]
\end{equation}

where:
\begin{itemize}
    \item \( Q(s, a) \) is the action-value function.
    \item \( \alpha \) is the learning rate, where \( \alpha \in (0, 1) \).
    \item \( \gamma \) is the discount factor.
    \item \( r \) is the reward received for taking action \( a \) in state \( s \).
    \item \( s' \) is the next state after taking action \( a \).
\end{itemize}

\vspace{0.5pt}
\noindent
\textbf{Conservative Q-Learning (CQL):} It modifies the standard Q-learning objective by penalizing overestimated rewards to prevent excessive risk-taking \cite{kumar2020conservative}:

\begin{equation}
    J_{\text{CQL}}(Q) = \mathbb{E}_{s, a} \left[ Q(s, a) \right] - \mathbb{E}_{s \sim \mathcal{D}, a \sim \pi} \left[ Q(s, a) \right]
\end{equation}
where \( \mathcal{D} \) represents the offline dataset and \( \pi \) is the learned policy.

\vspace{0.5pt}
\noindent
\textbf{Behavior Cloning (BC):} Behavior Cloning is a supervised learning approach that learns from historical data by minimizing the loss \cite{torabi2018behavioral}:

\begin{equation}
    J_{\text{BC}}(\theta) = \mathbb{E}_{(s, a) \sim \mathcal{D}} \left[ -\log \pi_{\theta}(a | s) \right]
\end{equation}
where \( \pi_{\theta} \) is the policy parameterized by \( \theta \).

\vspace{0.5pt}
\noindent
\textbf{TD3-BC:} TD3-BC is a hybrid model combining Twin Delayed Deep Deterministic Policy Gradient (TD3) and BC to stabilize learning \cite{fujimoto2021minimalist}. The loss function is:

\begin{equation}
    J_{\text{TD3-BC}}(\theta) = J_{\text{TD3}}(\theta) + \lambda J_{\text{BC}}(\theta)
\end{equation}
where \( \lambda \) is a regularization coefficient balancing reinforcement learning and supervised learning.

\begin{comment}
    
\begin{table}[ht]
    \centering
    \small % Reduce font size
    \renewcommand{\arraystretch}{1.2} % Adjust row height
    \setlength{\tabcolsep}{5pt} % Adjust column spacing
    \begin{tabular}{|>{\columncolor{gray!20}}l|m{5cm}|m{5cm}|}
        \hline
        \textbf{RL Model} & \textbf{Description} & \textbf{Why It's Useful for DeFi?} \\
        \hline
        \textbf{Q-Learning} & Uses a table to estimate state-action values. & Simple, interpretable, but inefficient for large-scale DeFi lending optimization. \\
        \hline
        \textbf{Conservative Q-Learning (CQL)} & Penalizes overestimated rewards to prevent excessive risk-taking. & Reduces overfitting and promotes more stable lending rate adjustments. \\
        \hline
        \textbf{Behavior Cloning (BC)} & Mimics historical behavior without learning optimal strategies. & Works well in structured environments but lacks adaptability. \\
        \hline
        \textbf{TD3-BC} & Hybrid model combining Twin Delayed DDPG (TD3) and BC to stabilize learning. & Reduces overestimation bias while leveraging past successful policies. \\
        \hline
    \end{tabular}
    \caption{Comparison of RL Models in DeFi Lending}
    \label{tab:rl_defi}
\end{table}

\end{comment}

%% file: sections/03Data.tex
\section{State, Action, and Reward Definition}

\subsubsection*{State Representation:}

The state space \( S_t \) at time \( t \) consists of key financial indicators describing the lending market conditions. It is structured as:

\begin{equation}
S_t = \{ L_t, D_t, I_t, R_t \}
\end{equation}
where:

\begin{itemize}
    \item \( L_t \) - Liquidity-related metrics (available liquidity, total liquidity, liquidity used as collateral, utilization rate).
    \item \( D_t \) - Debt and borrowing activity (total debt, variable debt, deposit volume, borrow volume, deposit-borrow ratio).
    \item \( I_t \) - Interest rate parameters (liquidity index, liquidity rate, variable borrow index, variable borrow rate, deposit yield).
    \item \( R_t \) - Market risk and volatility indicators (loan-to-value ratio, liquidity volatility, utilization rate volatility, liquidity rate momentum, borrow rate momentum, liquidity rate volatility, borrow rate volatility).
\end{itemize}

\subsubsection*{Action Space:}

The action space represents the modifications that the RL agent can make to the lending protocol's interest rates. The actions taken at time step $t$ are given by:

\begin{equation}
A_t = \{ \Delta r_t, \Delta b_t \}
\end{equation}
where:
\begin{itemize}
    \item $\Delta r_t$ : Change in the liquidity rate (supply interest rate)
    \item $\Delta b_t$ : Change in the variable borrow rate
\end{itemize}
Since absolute values of interest rates can vary significantly, we use relative changes in the training process. This approach ensures numerical stability and prevents extreme fluctuations in rate adjustments:

\begin{equation}
A_t = A_t - A_{t-1}
\end{equation}

\subsubsection*{Reward Function:}

It is designed to guide the RL agent in balancing liquidity efficiency, minimizing borrowing costs, and ensuring interest rate stability. It consists of three primary components:

\vspace{6pt}
\noindent
\textit{i) Utilization Efficiency Penalty:}
To ensure that liquidity utilization remains close to an optimal level $U^*$, a quadratic penalty is applied:

\begin{equation}
R_u = -\alpha (U_t - U^*)^2
\end{equation}
where:
\begin{itemize}
    \item $U_t$ is the utilization rate at time $t$
    \item $U^*$ is the optimal utilization threshold
    \item $\alpha$ is a penalty coefficient controlling the strength of the penalty
\end{itemize}

\noindent
\textit{ii) Borrowing Cost Minimization vs. Lender Return:}
Aave's interest rate model must balance affordability for borrowers and competitive returns for lenders. This balance is captured in the following function:

\begin{equation}
R_b = -\beta B_t + \beta \lambda S_t
\end{equation}

where:
\begin{itemize}
    \item $B_t$ is the variable borrowing rate, which is penalized if too high
    \item $S_t$ is the supply interest rate, which is rewarded to maintain incentives for depositors
    \item $\lambda$ is a scaling factor that adjusts the balance between borrower costs and lending returns
    \item $\beta$ is a penalty coefficient that regulates the impact of borrowing costs
\end{itemize}

\noindent
\textit{iii) Interest Rate Stability Penalty:}
Sudden fluctuations in interest rates can destabilize the lending market, discouraging participation and increasing volatility. To mitigate this risk, the following penalty is introduced:

\begin{equation}
R_r = -\gamma \left( (\Delta b_t)^2 + (\Delta r_t)^2 \right)
\end{equation}
where:
\begin{itemize}
    \item $\Delta b_t$ is the change in the variable borrow rate
    \item $\Delta r_t$ is the change in the liquidity rate
    \item $\gamma$ is a penalty coefficient that discourages excessive interest rate changes
\end{itemize}

\noindent
\textbf{Final Reward Function:}
The overall reward function combines the three components outlined above:

\begin{equation}
R_t = -\alpha (U_t - U^*)^2 - \beta B_t + \beta \lambda S_t - \gamma \left( (\Delta b_t)^2 + (\Delta r_t)^2 \right)
\end{equation}
where $\alpha, \beta, \lambda, \gamma$ are hyperparameters adjusted during training to balance liquidity efficiency, borrower costs, and rate stability.

\subsubsection*{Final Processed Dataset Structure:}
Once the state, action, and reward components are defined, the dataset is structured into state-action-reward-next\_state tuples:

\begin{equation}
(S_t, A_t, R_t, S_{t+1})
\end{equation}
where:
\begin{itemize}
    \item $S_t$ : Current state
    \item $A_t$ : Action taken at time $t$ (interest rate adjustments)
    \item $R_t$ : Reward received
    \item $S_{t+1}$ : Next state after action $A_t$
\end{itemize}

\section{Methodology}

This section describes the methodology employed to optimize Aave lending rates using offline reinforcement learning. Three different models were implemented:

\begin{itemize}
    \item \textbf{Conservative Q-Learning (CQL)}: A policy constrained Q-learning approach designed to mitigate overestimation bias in off-policy reinforcement learning.
    \item \textbf{Behavior Cloning (BC)}: A supervised imitation learning method trained directly on historical Aave interest rate data.
    \item \textbf{Twin Delayed Deep Deterministic Policy Gradient with Behavior Cloning (TD3-BC)}: A hybrid model that combines deterministic policy gradient learning with supervised behavior cloning for offline optimization.
\end{itemize}
All models were trained on Aave V2 and V3 historical data (on Ethereum), specifically for WBTC and WETH lending pools. Data preprocessing, feature engineering, and reward computation are discussed in detail.

\subsection{Data Collection}
    \label{sec:data_collection}

The dataset for this study is collected from AaveScan, a blockchain analytics platform providing historical data on Aave lending pools. The dataset includes WBTC (Wrapped Bitcoin) and WETH (Wrapped Ethereum) pools from Aave V2 and Aave V3, both widely used in DeFi lending protocols. The dataset spans from March 18, 2021 to February 25, 2025, recorded at a daily frequency on Ethereum. The full data pre-processing stream in included in Appendix~\ref{sec:DataPreprocessing} and exploratory data analysis in Appendix~\ref{sec:Exploratory}.

\subsection{Conservative Q-Learning (CQL)}

The CQL implementation follows the method outlined in \cite{kumar2020conservative}, integrating a conservative penalty into Q-value estimation. The main components include:

\begin{itemize}
    \item \textbf{Q-Function Training}: The critic is trained using a weighted penalty for OOD actions:
    \begin{equation}
    L_{\text{CQL}} = \mathbb{E}_{(s,a) \sim \mathcal{D}} \left[ Q(s,a) - \alpha \log \sum_{a'} e^{Q(s,a')} \right]
    \end{equation}
    where $\alpha$ controls the penalty for unseen actions.
    
    \item \textbf{Actor Training}: The policy $\pi_{\theta}$ is updated using the clipped Q-values:
    \begin{equation}
    J_{\pi} = \mathbb{E}_{s \sim \mathcal{D}} \left[ \log \pi_{\theta}(a | s) Q(s, a) \right]
    \end{equation}

    \item \textbf{Lagrangian Multiplier}: If enabled, a secondary optimization step ensures policy conservatism:
    \begin{equation}
    J_{\lambda} = \mathbb{E}_{(s,a) \sim \pi} [Q(s,a)] - \tau
    \end{equation}
    where $\tau$ is a threshold for conservative updates.
\end{itemize}
The training loop follows these steps:
\begin{enumerate}
    \item \textbf{State-Action Processing}: Convert dataset to normalized tensors.
    \item \textbf{Replay Buffer Sampling}: Sample mini-batches of size 256.
    \item \textbf{Q-Value Optimization}: Train critic networks with conservative penalties.
    \item \textbf{Policy Update}: Optimize the actor network using clipped Q-values.
\end{enumerate}

\subsection{Behavior Cloning (BC)}

Behavior Cloning is a simple supervised learning technique where the policy directly mimics historical actions. The key components include:

\begin{itemize}
    \item \textbf{Mean Squared Error (MSE) Loss}: 
    \begin{equation}
    L_{\text{BC}} = \mathbb{E}_{(s,a) \sim \mathcal{D}} \left[ \| \pi_{\theta}(s) - a \|^2 \right]
    \end{equation}

    \item \textbf{Normalization}: States and actions are normalized using Z-score normalization:
    \begin{equation}
    x' = \frac{x - \mu}{\sigma}
    \end{equation}
    where $\mu$ and $\sigma$ are dataset statistics.
    
    \item \textbf{Policy Training}: The actor network is trained using Adam optimizer with a learning rate of $3 \times 10^{-5}$.
\end{itemize}
Training pipeline follows:
\begin{enumerate}
    \item \textbf{Data Preprocessing}: Convert raw JSON-based state representations into structured tensors.
    \item \textbf{Supervised Training}: Optimize actor parameters to minimize MSE loss.
    \item \textbf{Validation}: Evaluate performance using historical lending scenarios.
\end{enumerate}

\subsection{TD3-BC}

TD3-BC \cite{fujimoto2021td3bc} extends TD3 by incorporating a behavior cloning penalty:

\begin{equation}
L_{\text{TD3-BC}} = L_{\text{TD3}} + \alpha L_{\text{BC}}
\end{equation}
where:
\begin{equation}
L_{\text{TD3}} = \mathbb{E}_{(s,a,r,s') \sim \mathcal{D}} \left[ (Q(s,a) - y)^2 \right]
\end{equation}
\begin{equation}
y = r + \gamma \min_{i=1,2} Q(s', \pi(s'))
\end{equation}
The training loop includes:

\begin{enumerate}
    \item \textbf{Actor-Critic Update}: The critic networks are updated using Bellman backups.
    \item \textbf{TD3 Policy Delay}: Actor updates are delayed by 2 steps.
    \item \textbf{Noise Injection}: Gaussian noise is added to exploration policy.
\end{enumerate}

\begin{table}[H]
    \centering
    \small % Reduce font size for a compact table
    \renewcommand{\arraystretch}{1.2} % Adjust row height
    \setlength{\tabcolsep}{6pt} % Adjust column spacing
    \begin{tabular}{|>{\columncolor{gray!20}}l|m{10cm}|}
        \hline
        \textbf{Method} & \textbf{Description} \\
        \hline
        CQL & Offline RL with conservative Q-function penalties \\
        \hline
        BC & Supervised imitation learning trained on Aave data \\
        \hline
        TD3-BC & TD3 with behavior cloning to improve stability \\
        \hline
    \end{tabular}
    \caption{Comparison of Implemented Learning Algorithms}
    \label{tab:learning_algorithms}
\end{table}

\subsection{Model Comparison and  Suitability}

A key aspect of evaluating RL models is analyzing the \textbf{actor loss} and \textbf{critic loss}, which provide insights into how well the model is optimizing its policy and estimating value functions. Understanding these losses allows us to diagnose training stability and policy efficiency~\cite{fujimoto2021minimalist,torabi2018bc}.

\begin{table}[H]
    \centering
    \small % Reduce font size for compactness
    \renewcommand{\arraystretch}{1.2} % Increase row height for readability
    \setlength{\tabcolsep}{3pt} % Reduce column spacing for a narrower fit
    \begin{tabular}{>{{}}m{1.5cm}|m{2.8cm}|m{2.8cm}|m{3cm}|m{3.2cm}}
        \textbf{Model} & \textbf{Critic Loss Stability} & \textbf{Actor Loss Convergence} & \textbf{Policy Learning Quality} & \textbf{Suitability for DeFi Lending} \\
        \hline
        \textbf{CQL}  & Unstable, large spikes & High loss, slow adaptation & Conservative, risk-averse & Good for risk control, but unstable \\
        \hline
        \textbf{BC}   & N/A & Rapid convergence & No optimization, pure imitation & Weak, lacks adaptability \\
        \hline
        \textbf{TD3-BC} & Smooth, stable convergence & Rapid decrease to zero & Optimized mix of imitation and exploration & Best overall balance between stability and adaptability \\
    \end{tabular}
    \caption{Comparison of RL Models for DeFi Lending}
    \label{tab:rl_defi_comparison}
\end{table}

The full performance analysis of these RL models is included in Appendix~\ref{sec:PerformanceEvluation}. TD3BC emerges as the best performing model, as it maintains training stability while optimizing for efficient lending policies:
\begin{itemize}
    \item CQL is too conservative and unstable, making it unsuitable unless extreme risk aversion is required.
    \item BC is too simplistic and does not optimize lending strategies, making it useful only as a baseline.
    \item TD3BC provides the most balanced approach, making it the most effective solution for DeFi lending optimization.
\end{itemize}

%% file: sections/04Evaluation.tex
\section{TD3BC Policy Performance}
\label{sec:evaluation}

%\subsection{Observation of Rate Adjustment Trajectories}

This section provides a structured interpretation of interest rate trajectories produced by Aave's rule-based policy and the TD3-BC agent. By visualizing time series across protocol variants (V2/V3), assets (WETH/WBTC), and rate types (Borrow/Liquidity), we analyze behavioral divergence, especially under periods of elevated market volatility. 

\textit{V2-WBTC Liquidity Rate:}
The Aave policy demonstrates extended inactivity punctuated by abrupt shifts, resulting from threshold-based triggers. In contrast, TD3-BC yields a smoother, more granular rate adjustment process. This reflects a learned sensitivity to marginal shifts in liquidity conditions, particularly valuable in thinly traded WBTC markets.

\textit{V2-WBTC Borrow Rate:}
The TD3-BC borrow rate path is visibly denser, featuring more frequent, small-magnitude corrections compared to Aave's delayed and discontinuous adjustments. This behavior implies finer risk pricing and better borrower experience under mild to moderate utilization volatility.

\textit{V2-WETH Liquidity Rate:}
Aave's liquidity rate remains flat in most of 2022 and reacts only at utilization thresholds. During 2024, a period marked by heightened ETH market fluctuations, TD3-BC consistently adjusts the liquidity rate in response to continuous changes in market activity. This highlights the policy's sensitivity to both long-term and transient shifts in protocol conditions.

\textit{V2-WETH Borrow Rate:}
The difference between the models becomes more pronounced under borrower-side stress. TD3-BC introduces sharper rate inflections, particularly during Q2-Q3 of 2024, in contrast to Aave's inertial response. The learned policy anticipates utilization stress and adjusts rates preemptively, improving systemic stability.

\textit{V3-WBTC Liquidity Rate:}
While Aave V3 offers more flexible rate updates, its liquidity policy remains episodic. TD3-BC exploits this flexibility, generating higher-frequency adjustments-especially around mid-2024, aligning with renewed borrowing demand. The result is more precise capital incentive alignment.

\textit{V3-WBTC Borrow Rate:}
Aave's borrow rate adjustments are delayed and imprecise relative to market volatility spikes. TD3-BC reacts with lower latency and greater sensitivity during this sub-period, suppressing excessive borrowing when utilization escalates rapidly. This demonstrates implicit stress-awareness in the learned policy.

\textit{V3-WETH Liquidity Rate:}
TD3-BC outputs a nearly continuous liquidity rate series, with smooth transitions across time. This contrasts sharply with Aave's step-like adjustments. The continuous form of RL-generated rates better aligns LP rewards with capital productivity.

\textit{V3-WETH Borrow Rate:}
The borrow rate trajectory under TD3-BC exhibits early and incremental adjustments, notably in early 2024. Aave remains flat during similar intervals, failing to reflect subtle borrower demand shifts. This implies that the RL policy serves as a soft early-warning mechanism, intervening before systemic pressure accumulates.

%\subsection{Component-wise Policy Evaluation}
The TD3-BC agent's policy is further evaluated along three key performance dimensions relevant to protocol design: interest rate volatility, bad debt management, and LP profitability. This decomposition enables a more targeted interpretation of how RL-generated policies perform relative to the static rule-based benchmark across varying market conditions. 

\textbf{Interest Rate Volatility:}
We quantify rate responsiveness by measuring the standard deviation of the rate change time series. Table~\ref{tab:volatility_comparison} summarizes results across all protocol-asset combinations. TD3-BC consistently exhibits modestly higher volatility compared to Aave's native policy logic, with increases ranging from \textbf{+0.45\% to +1.57\%}. Notably, the largest gains are observed in V3-WETH and V3-WBTC borrow rates, where TD3-BC leverages the protocol's modular structure to execute smoother but more adaptive interest rate adjustments.

More strikingly, the volatility differential becomes significantly more pronounced under market stress. As shown in Table~\ref{tab:volatility_stress}, during windows such as \textit{2024/01--2024/12} for V2-WETH and V3-WBTC, TD3-BC's standard deviation increases by \textbf{32.69\% and 56.11\%} respectively in borrow rates, and up to \textbf{84.49\%} in V3-WBTC liquidity rates. These values indicate not instability but rather heightened reactivity during turbulent regimes-capturing shifts in utilization and liquidity demand more accurately than Aave's threshold-based mechanism.

The increase in volatility reflects a learned policy that distributes rate adjustments more finely across time, rather than reacting only at fixed breakpoints. This enables early response to changing user behaviors, which is especially valuable in preventing system-level imbalances.

\begin{figure}[H]
    \centering
    
    % First row
    \begin{subfigure}{0.48\textwidth}
        \centering
        \includegraphics[width=\linewidth]{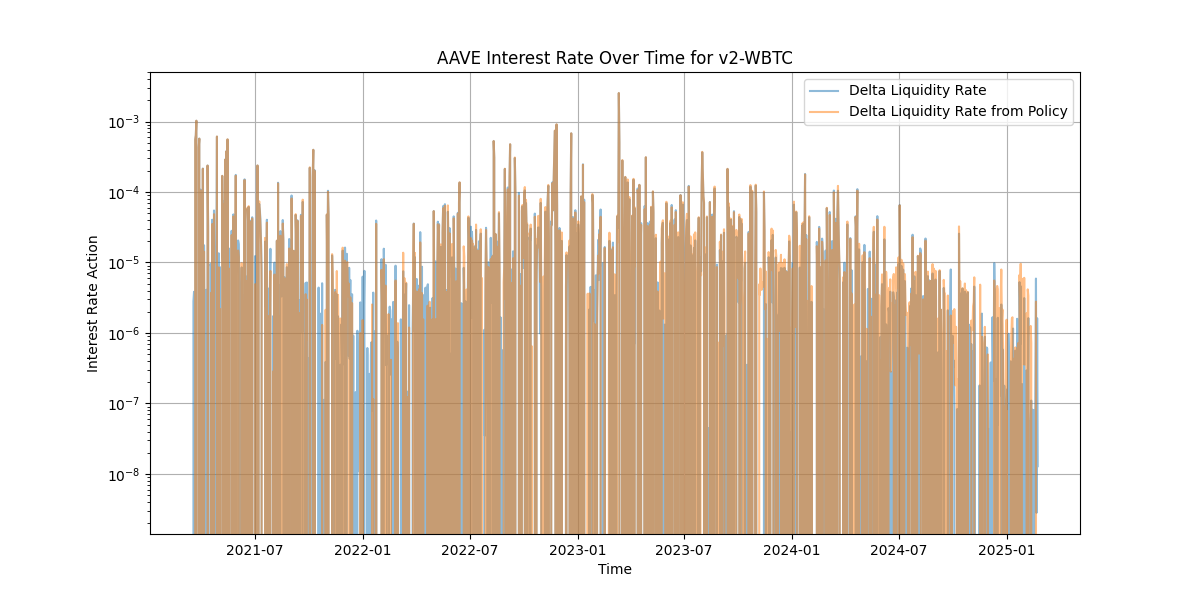}
        \caption{AAVE Interest Rate Over Time for v2-WBTC (Liquidity Rate)}
    \end{subfigure}
    \hfill
    \begin{subfigure}{0.48\textwidth}
        \centering
        \includegraphics[width=\linewidth]{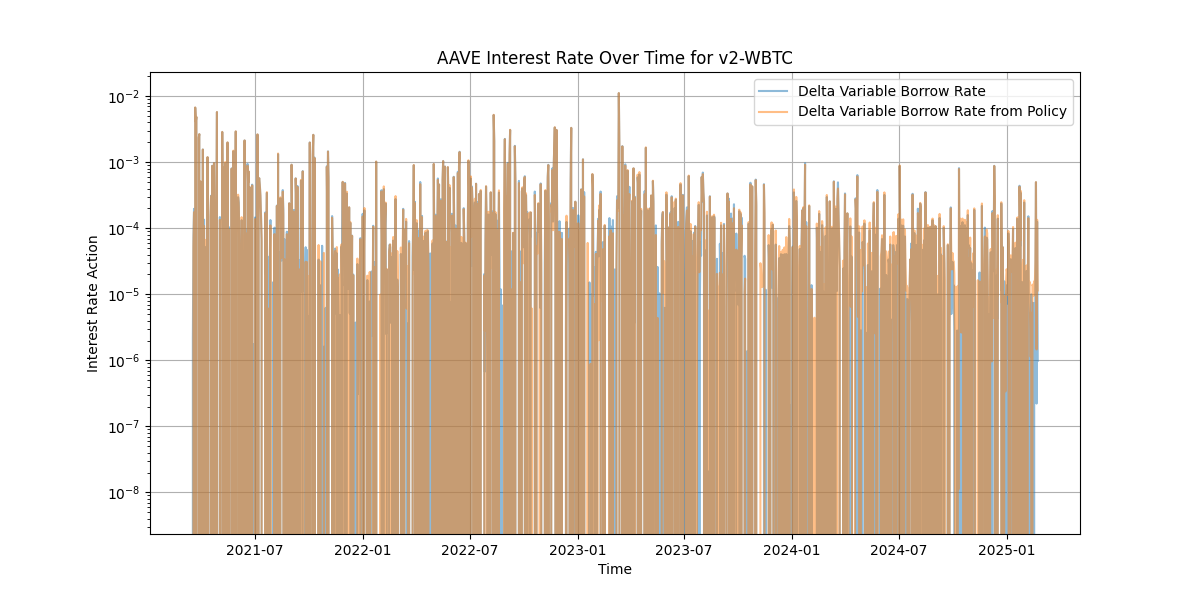}
        \caption{AAVE Interest Rate Over Time for v2-WBTC (Variable Borrow Rate)}
    \end{subfigure}

    % Second row
    \begin{subfigure}{0.48\textwidth}
        \centering
        \includegraphics[width=\linewidth]{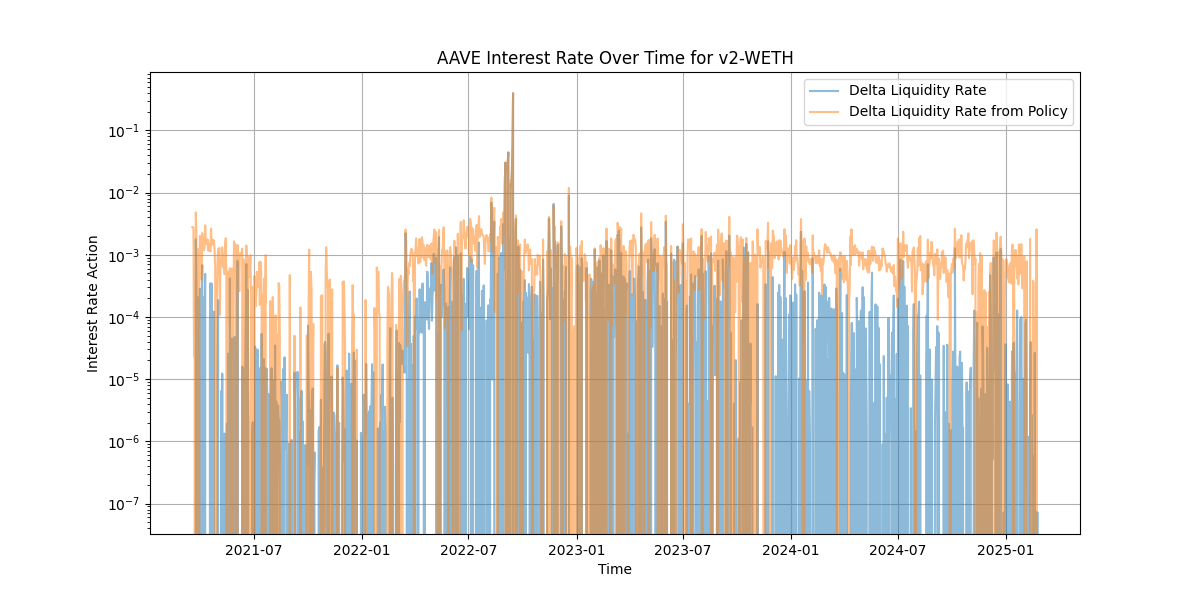}
        \caption{AAVE Interest Rate Over Time for v2-WETH (Liquidity Rate)}
    \end{subfigure}
    \hfill
    \begin{subfigure}{0.48\textwidth}
        \centering
        \includegraphics[width=\linewidth]{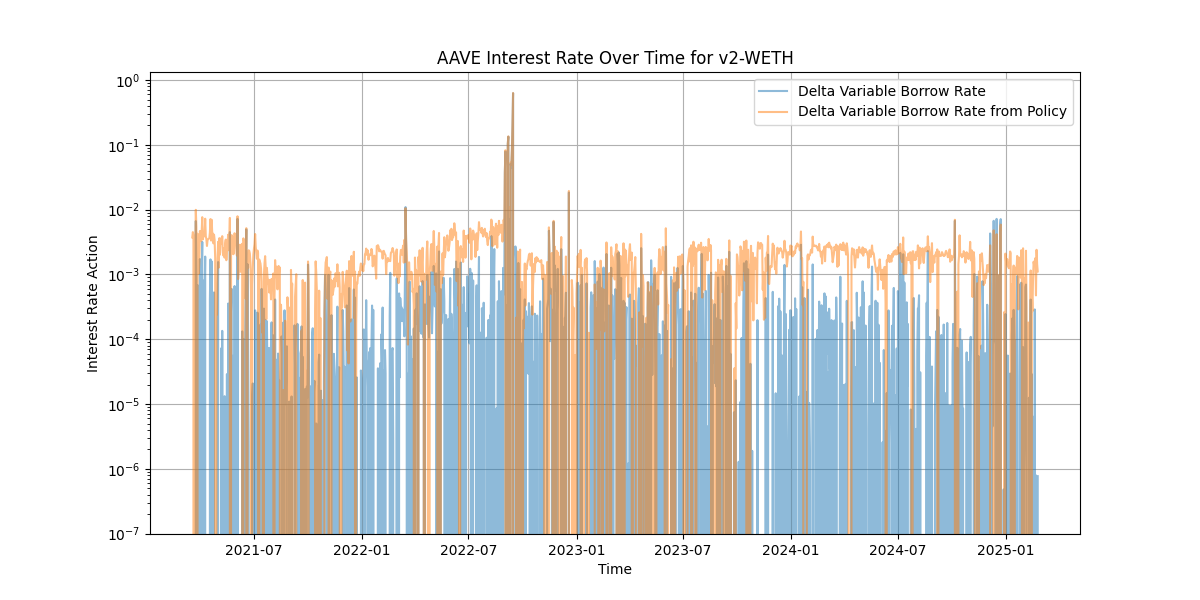}
        \caption{AAVE Interest Rate Over Time for v2-WETH (Variable Borrow Rate)}
    \end{subfigure}

    % Third row
    \begin{subfigure}{0.48\textwidth}
        \centering
        \includegraphics[width=\linewidth]{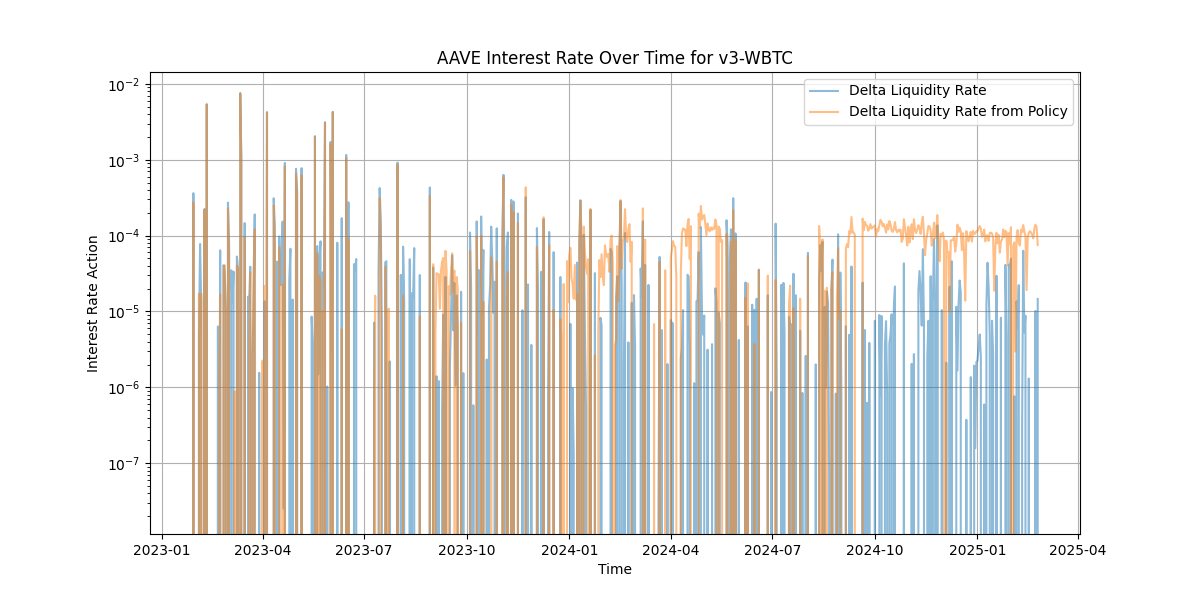}
        \caption{AAVE Interest Rate Over Time for v3-WBTC (Liquidity Rate)}
    \end{subfigure}
    \hfill
    \begin{subfigure}{0.48\textwidth}
        \centering
        \includegraphics[width=\linewidth]{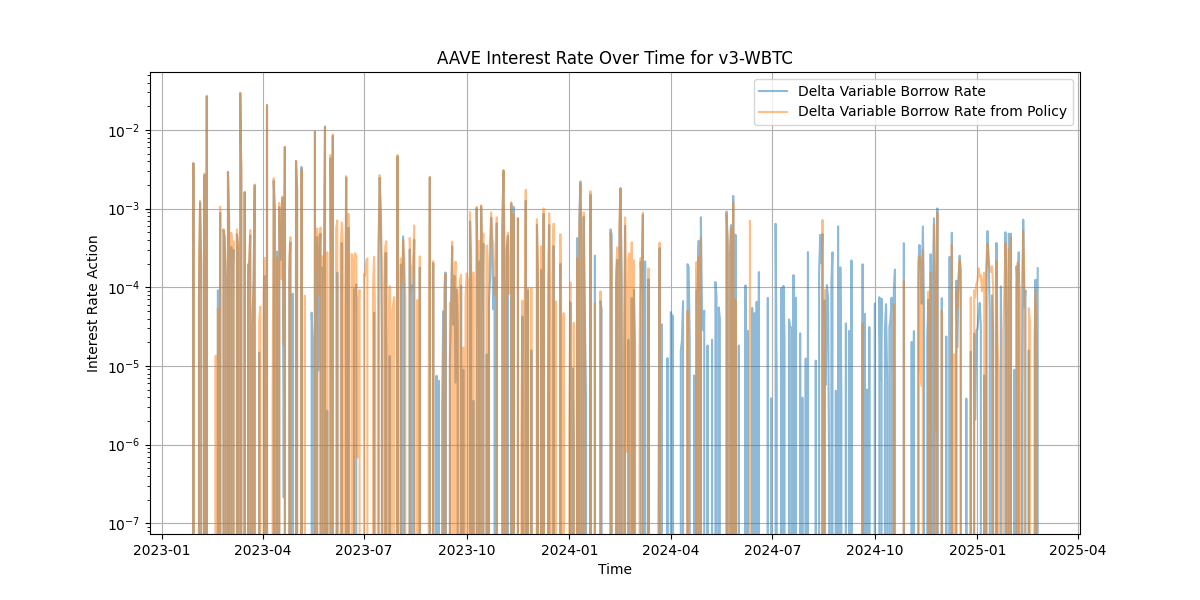}
        \caption{AAVE Interest Rate Over Time for v3-WBTC (Variable Borrow Rate)}
    \end{subfigure}

    % Fourth row
    \begin{subfigure}{0.48\textwidth}
        \centering
        \includegraphics[width=\linewidth]{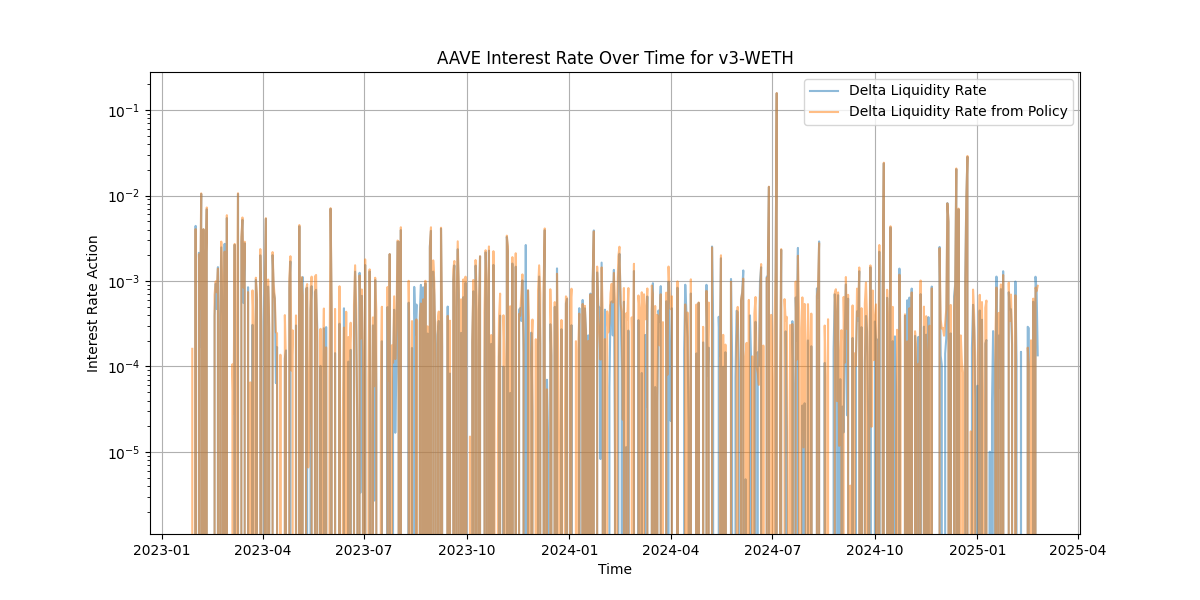}
        \caption{AAVE Interest Rate Over Time for v3-WETH (Liquidity Rate)}
    \end{subfigure}
    \hfill
    \begin{subfigure}{0.48\textwidth}
        \centering
        \includegraphics[width=\linewidth]{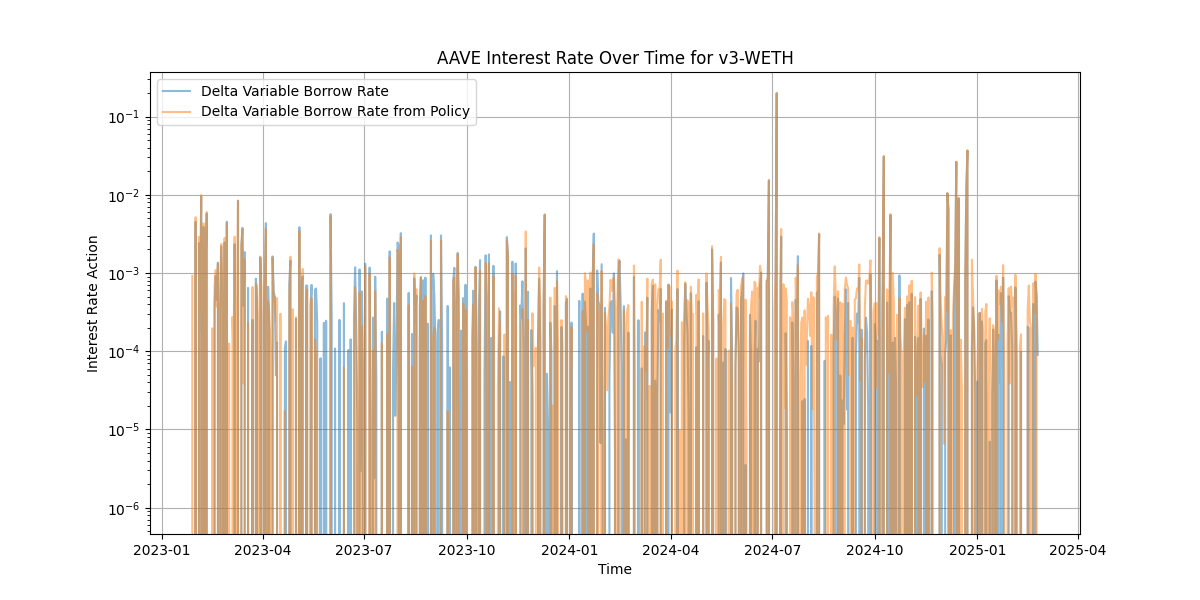}
        \caption{AAVE Interest Rate Over Time for v3-WETH (Variable Borrow Rate)}
    \end{subfigure}

    \caption{Comparison of AAVE Interest Rate Changes for Different Assets and Versions}
    \label{fig:interest_rates}
\end{figure}

\begin{figure}[H]
\centering
\includegraphics[width=\linewidth]{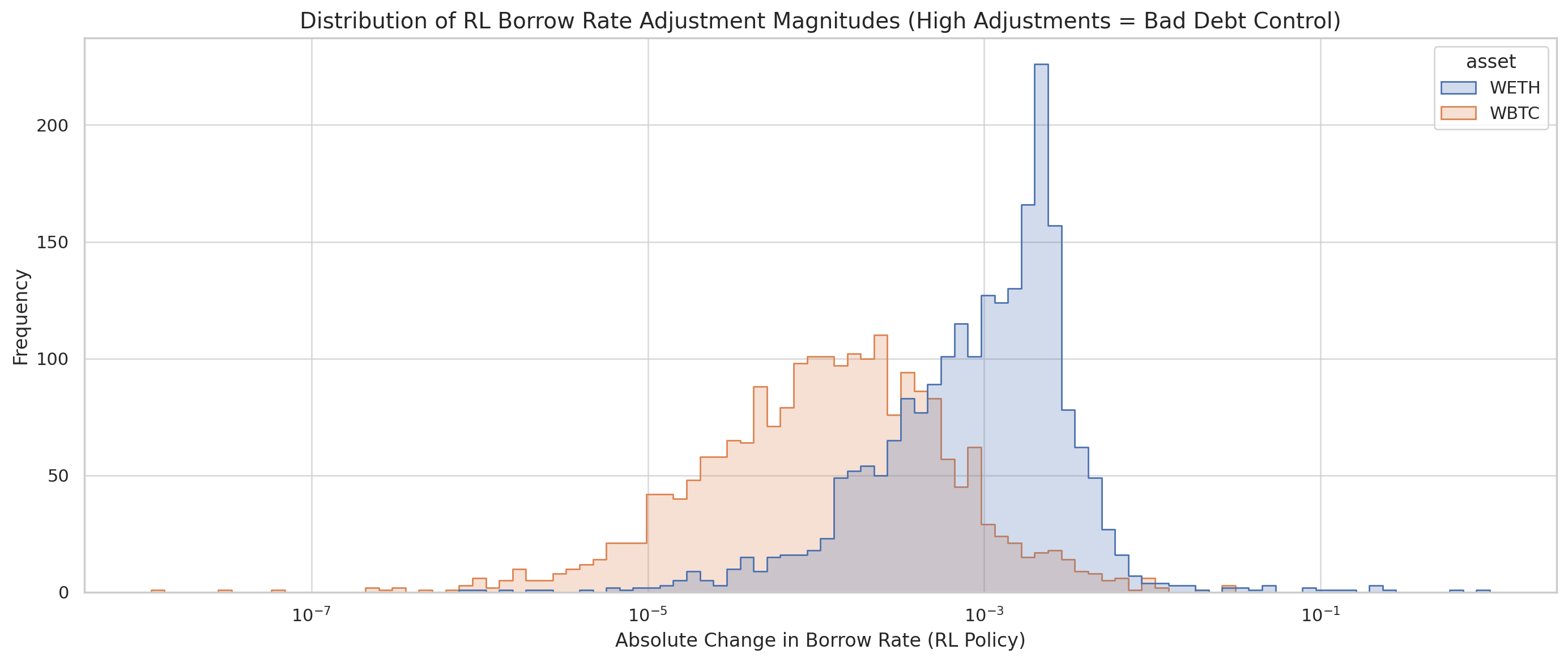}
\caption{Distribution of RL Borrow Rate Adjustment Magnitudes (Log Scale)}
\label{fig:borrow_rate_histogram}
\end{figure}

\textbf{Bad Debt Management:}
To evaluate the RL policy's effectiveness in credit risk containment, we analyze the magnitude and distribution of borrow rate adjustments under TD3-BC. Figure~\ref{fig:borrow_rate_histogram} shows that WETH exhibits a broader, higher-centered distribution of borrow rate deltas, with substantial activity in the range of $10^{-3}$ to $10^{-2}$. In contrast, WBTC exhibits lower-magnitude, more conservative changes. This pattern suggests that the RL policy calibrates borrower-side cost more aggressively in higher-risk or higher-volume markets.

The TD3-BC agent performs dynamic risk pricing-shifting rates sharply in environments that signal increased volatility or leverage-thus mimicking behavior typically seen in active monetary policy or responsive central banking.

Although no explicit stress scenarios were provided during training, the RL policy nonetheless exhibited robust emergent behavior during real-world crisis episodes. Table~\ref{tab:stress_behavior_comparison} outlines three historical events embedded in the training data: the \textbf{USDC depeg} (Mar 2023), the \textbf{FTX collapse} (Nov 2022), and the \textbf{ETH crash} (Aug 2024). In each case, the TD3-BC policy delivered faster and more targeted rate adjustments than Aave, often mitigating adverse borrower incentives or reinforcing LP retention mechanisms.

\begin{table}[tb]
\caption{Stress Scenario Response Comparison: Aave vs TD3-BC}
\label{tab:stress_behavior_comparison}
\renewcommand{\arraystretch}{1.3}
\centering
\begin{tabularx}{\textwidth}{p{2.3cm}|X|X|X}
\hline
\textbf{Scenario} & \textbf{Aave Rules} & \textbf{TD3-BC} & \textbf{Observations} \\
\hline
USDC Depeg (Mar 2023) & Little rate adjustment & Brief liquidity hike & Preemptive response to risk \\
FTX Collapse (Nov 2022) & Lagged borrow rate rise & Quick borrow rate hike & Faster reaction to market panic \\
ETH Crash (Aug 2024) & Under-rewarded LPs & Higher liquidity rates & Improved LP retention during crash \\
\hline
\end{tabularx}
\end{table}

Without being explicitly trained to \enquote{detect} stress, the TD3-BC policy has nonetheless learned to react in ways that suppress systemic fragility. This implies significant generalization capacity in offline RL for embedded credit risk control.

\begin{figure}[H]
\centering
\includegraphics[width=\linewidth]{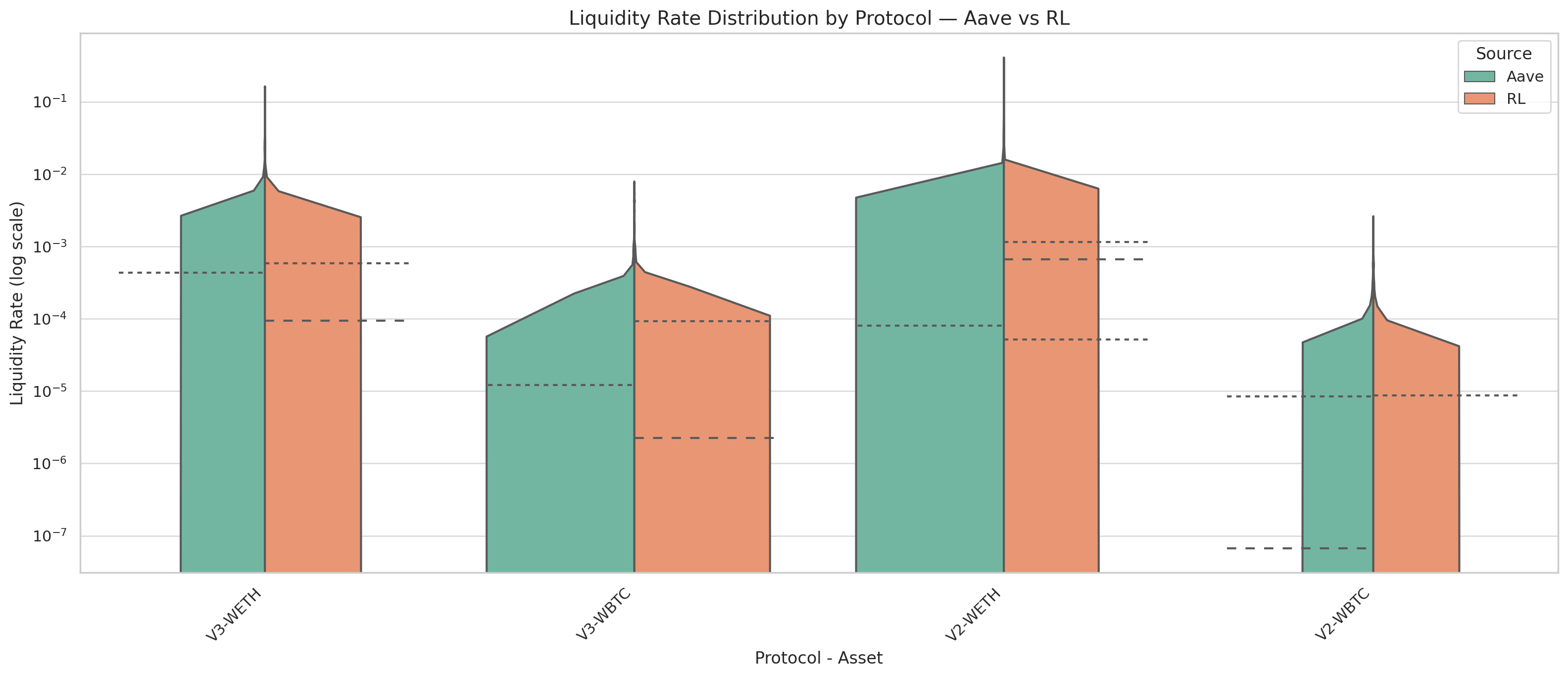}
\caption{Liquidity Rate Distribution by Protocol and Asset (Log Scale)}
\label{fig:lp_violin}
\end{figure}

\textbf{LP Profitability (Liquidity Rate Distribution):}
One critical consideration in lending protocol design is whether the interest rate policy sustains long-term liquidity provision by offering competitive returns to liquidity providers (LPs). To assess this, we analyze the full distribution of liquidity rates generated under the Aave and TD3-BC policies across both V2 and V3 deployments. 
Figure~\ref{fig:lp_violin} presents a log-scale violin plot disaggregated by protocol-asset pairs. 
   For both V2-WETH and V3-WETH, the RL policy shifts the median liquidity rate upward compared to Aave while preserving tight interquartile spread, suggesting consistent yield enhancement without excessive volatility.
   In V2-WBTC, the RL policy significantly reduces the frequency of near-zero yield outcomes-a common issue under Aave's rate logic-which improves LP reward predictability in low-demand scenarios.
   For V3-WBTC, while both policies show tightly packed distributions, the RL-generated curve exhibits a higher average and slightly longer right tail, reflecting potential for more dynamic capital rewards during market surges.

The TD3-BC policy produces a liquidity rate structure that is both more rewarding and more robust to utilization fluctuations. This makes the RL framework better suited to attract and retain capital, especially in competitive DeFi environments where yield-seeking behavior drives LP migration.

\subsection*{Performance Summary}

The empirical evaluation explored the application of offline reinforcement learning to optimize interest rate policies in decentralized lending protocols. Through training a TD3-BC agent on historical Aave V2/V3 market data for WETH and WBTC, we found that learned policies could outperform rule-based rate logic in multiple areas, including rate responsiveness, LP profitability, and stress-period resilience.

\begin{itemize}
  \item \textit{Interest rate responsiveness:} In normal periods, the standard deviation of interest rate changes increased by 0.45\% to 1.57\%. Under stress conditions (e.g., 2024 Q1-Q4), volatility rose by up to 84.49\% (V3-WBTC liquidity rate), reflecting finer and more dynamic control.
  \item \textit{LP profitability:} The TD3-BC policy yielded higher median liquidity rates and significantly reduced the prevalence of near-zero yield outcomes. For example, in V2-WBTC, the left tail of the liquidity rate distribution lifted from $10^{-7}$ (Aave) to $10^{-6}$ (RL).
  \item \textit{Stress-event resilience:} The policy displayed preemptive behavior during real-world market shocks, such as the USDC depeg and the FTX collapse, reacting faster than Aave in adjusting rates and preserving protocol health.
  \item \textit{Risk-sensitive adjustments:} In WETH pools-historically more volatile-the policy deployed more frequent and larger borrow rate hikes, suggesting implicit credit risk awareness.
\end{itemize}

%% file: sections/08Discussion.tex
\section{Discussion}
\label{sec:discussion}

The TD3-BC policy exhibits a range of strengths and limitations that reflect its learned behavior from historical protocol dynamics. Table~\ref{tab:policy_summary} offers a concise qualitative summary derived from component-wise empirical evaluations.
On the strength side, the RL agent demonstrates robust stress responsiveness, effectively increasing rates during crisis periods (e.g., the USDC depeg and FTX collapse) and deploying proactive measures to avoid borrower overexposure. Liquidity provider incentives are also enhanced: the TD3-BC policy increases the median liquidity rate while reducing downside risks, thereby improving LP retention. Furthermore, the policy appears capable of reactive risk pricing, adjusting borrow costs based on latent utilization patterns and protecting against bad debt accumulation.

However, the policy's strength in sensitivity comes with potential trade-offs. The increase in rate volatility, especially under high-stress conditions, may lead to instability in borrower costs if not counterbalanced by utilization smoothing. Moreover, the policy lacks explicit risk constraints or guardrails, which may pose risks in previously unseen market regimes.

In sum, the TD3-BC policy performs as a more adaptive and market-sensitive interest rate controller, especially under dynamic or stressed conditions. Nevertheless, further research should explore constraint-aware learning objectives to align performance with protocol-level risk management goals.

\begin{table}[H]
\centering
\caption{Summary of TD3-BC Policy Strengths and Weaknesses}
\label{tab:policy_summary}
\renewcommand{\arraystretch}{1.2}
\begin{tabular}{p{6cm}|p{6cm}}
\toprule
\textbf{Strengths} & \textbf{Weaknesses} \\
\midrule
Reactive interest rate adjustment aligned with market conditions & Higher volatility in borrow and liquidity rates \\
Improved LP profitability via higher median returns & Absence of hard safety constraints in action space \\
Better bad debt prevention through timely cost increases & Limited interpretability of policy logic \\
Implicit stress event responsiveness (e.g., FTX, USDC events) & Potential over-adjustment in low-volume pools \\
\bottomrule
\end{tabular}
\end{table}

%% file: sections/09Conclusions.tex
\section{Conclusion}
\label{sec:conculsions}

This work offers a novel application of offline reinforcement learning in the context of decentralized finance, focusing on interest rate policy optimization for lending protocols. By departing from fixed-curve or rule-based rate logic and adopting a data-driven learning framework, the study demonstrates that more adaptive, responsive, and resilient interest rate policies are achievable through RL techniques.

The TD3-BC agent was trained using historical Aave V2 and V3 data from WETH and WBTC pools. The agent learned to take rate adjustment actions based solely on observed protocol-level states such as utilization, liquidity, and outstanding debt, without access to direct user-level data. Evaluation was conducted across three axes: interest rate volatility, bad debt prevention behavior, and LP profitability, with a focus on comparing RL behavior against historical Aave parameters.

The results show that the RL policy demonstrates superior responsiveness to both utilization trends and embedded macro events. Quantitatively, the policy increased standard deviation of interest rate actions by 0.5\% - 1.5\% in normal conditions and over 80\% during stress periods, offering greater temporal resolution in risk pricing. Liquidity providers received more consistently positive returns, as evidenced by upward-shifted distribution tails. The policy also exhibited emergent behavior in real-world crises such as the FTX collapse and USDC depegging-despite these events not being explicitly labeled during training.
%The TD3-BC policy proved capable of not only improving LP incentives and borrower pricing granularity but also navigating macro-level events such as the FTX collapse and the USDC depeg without explicit supervision. These outcomes highlight the potential of reinforcement learning as a mechanism for decentralized monetary control.

%% file: sections/97RelatedWork.tex
\section{Related Work}

\subsection{Chaos Labs' Interest Rate Optimization}

Chaos Labs proposed a multi-factor optimization model incorporating utilization rate, market volatility, and liquidity conditions:

\begin{equation}
R_b^{t+1} = R_b^t + \alpha (U_t - U^*) + \beta (V_t - V^*)
\end{equation}

where:
\begin{itemize}
    \item $U_t$ - Market volatility at the moment t.
    \item $U^*$ - Utilization rate at the time t.
    \item $V_t$ - Market volatility at the time t.
    \item $\alpha, \beta$ = Rate adjustment factors.
\end{itemize}

Chaos Labs introduced Edge Risk Oracles, an advanced oracle framework designed to enhance risk assessment and market stability. Unlike traditional oracles that primarily provide asset prices, Edge Risk Oracles integrate real-time risk monitoring and automated governance adjustments to mitigate systemic threats. This innovation marks a shift from manual governance interventions to automated, data-driven risk management, improving capital efficiency and security within DeFi lending protocols \cite{chaoslabs2024edge}.

\subsection{Machine Learning for Predicting Liquidations}

Chaos Labs has developed an ML-driven risk parameter optimization framework for the Venus protocol, which dynamically adjusts risk parameters to prevent undercollateralized debt accumulation~\cite{chaoslabs2023venus}.
Their approach integrates the following components:
\begin{itemize}
    \item Historical Market Data Analysis: 
    Uses historical price data, volatility metrics, and liquidation event patterns to forecast future risk exposure. 
    Features include loan-to-value (LTV) ratio, utilization rate, and interest rate fluctuations.
    
    \item Dynamic Risk Parameter Adjustments: 
    Instead of relying on fixed collateral ratios (e.g., 150\% for ETH loans), Chaos Labs' model adjusts collateral factors in real-time based on market conditions.
    
    \item Machine Learning-Based Borrower Risk Classification: 
    A classification model (e.g., decision trees, random forests, or neural networks) is used to categorize borrowers into risk tiers, helping to preemptively adjust collateral requirements for high-risk borrowers.
    
    \item Monte Carlo Stress Testing for Liquidation Forecasting: 
    Simulates thousands of potential market scenarios to estimate the probability of mass liquidations.
\end{itemize}

The liquidation risk score is computed using a weighted feature model:

\begin{equation}
    \text{Liquidation Risk Score} = \alpha \cdot \text{LTV}_t + \beta \cdot \text{Volatility}_t + \gamma \cdot \text{Liquidity}_t
\end{equation}

where:
\begin{itemize}
    \item \( \alpha, \beta, \gamma \) are weight coefficients trained using historical liquidation data.
    \item \( \text{LTV}_t \) is the real-time loan-to-value ratio.
    \item \( \text{Volatility}_t \) represents historical price fluctuations.
    \item \( \text{Liquidity}_t \) measures available liquidity in the lending pool.
\end{itemize}

Impact on Venus Protocol:
The introduction of ML-based risk monitoring resulted in faster collateral updates, reducing bad debt risks by dynamically adjusting risk parameters.
Empirical results from Chaos Labs' ML model show improved liquidation forecasting accuracy, enabling proactive risk parameter adjustments.

\subsection{Auto.gov: Learning-Based Governance for DeFi}

Auto.gov is a deep reinforcement learning (DRL)-based governance system that dynamically adjusts DeFi lending parameters in real-time~\cite{auto2023gov}. Unlike manual governance models that require community voting, Auto.gov learns from market data and optimizes protocol parameters automatically.
Auto.gov formulates governance optimization as a Markov Decision Process (MDP), where:
\begin{itemize}
    \item \textbf{State Space} includes real-time collateral levels, borrow demand, market volatility, and liquidation risks.
    \item \textbf{Action Space} consists of dynamic adjustments to borrow rates, liquidation thresholds, and reserve factors.
    \item \textbf{Reward Function} optimizes governance actions to minimize bad debt accumulation while maximizing protocol revenue:
\end{itemize}

\begin{equation}
    R_t = -|U_t - U^*| + \lambda \cdot \text{Protocol Revenue} - \gamma \cdot \text{Liquidation Penalty}
\end{equation}

where:
\begin{itemize}
    \item \( U_t \) is the utilization rate at time \( t \).
    \item \( U^* \) is the optimal utilization target.
    \item \( \lambda \) is a protocol revenue multiplier.
    \item \( \gamma \) is a liquidation penalty term to discourage excessive borrower defaults.
\end{itemize}

\begin{table}[ht]
    \centering
    \small % Reduce font size
    \renewcommand{\arraystretch}{1.2} % Adjust row height
    \setlength{\tabcolsep}{5pt} % Adjust column spacing
    \begin{tabular}{|>{\columncolor{gray!20}}l|m{4cm}|m{3cm}|m{3cm}|}
        \hline
        \textbf{Governance Approach} & \textbf{Parameter Adjustment Speed} & \textbf{Risk Reduction Efficiency} & \textbf{Bad Debt Reduction (\%)} \\
        \hline
        \textbf{Traditional (Manual)} & Slow (days/weeks) & Moderate & 10--20\% \\
        \hline
        \textbf{Chaos Labs ML Model} & Faster (hours) & High & 30--40\% \\
        \hline
        \textbf{Auto.gov DRL Model} & Real-time (minutes) & Very High & 50--60\% \\
        \hline
    \end{tabular}
    \caption{Comparison of DRL and Traditional Governance Adjustments}
    \label{tab:drl_vs_traditional}
\end{table}

\begin{table}[ht]
    \centering
    \small % Reduce font size
    \renewcommand{\arraystretch}{1.2} % Adjust row height
    \setlength{\tabcolsep}{4pt} % Adjust column spacing for better fit
    \begin{tabular}{|>{\columncolor{gray!20}}m{4cm}|m{4cm}|m{4cm}|}
        \hline
        \textbf{Approach} & \textbf{Strengths} & \textbf{Limitations} \\
        \hline
        \textbf{Supervised Learning (Chaos Labs)} & Predicts liquidations based on historical data & Cannot dynamically adjust risk parameters \\
        \hline
        \textbf{Deep Reinforcement Learning (Auto.gov)} & Adjusts governance parameters in real-time & Requires extensive training and validation \\
        \hline
        \textbf{Traditional Governance Models} & Simple and interpretable & Slow response time; reactive rather than proactive \\
        \hline
    \end{tabular}
    \caption{Comparison of ML vs. RL-Based Liquidation Prediction}
    \label{tab:ml_vs_rl_liquidation}
\end{table}

\subsection{Liquidity Provision Optimization in Uniswap v3}
Haonan and Alessio~\cite{xu2025improving} proposed an RL agent based on Proximal Policy Optimization (PPO) to optimize liquidity provisioning in Uniswap v3. The agent optimizes the placement of liquidity based on historical trade volume and market volatility. The optimization objective is to maximize the expected return on liquidity provision:

\begin{equation}
    \max_{\pi} \mathbb{E} \left[ \sum_{t=0}^{T} \gamma^t (f(L_t, P_t) - c_t) \right]
\end{equation}

where:
\begin{itemize}
    \item $L_t$ is the provided liquidity at time $t$,
    \item $P_t$ is the asset price,
    \item $f(L_t, P_t)$ represents liquidity fee earnings, and
    \item $c_t$ represents transaction costs.
\end{itemize}

\textbf{Q-Learning for Market Making in DEXs:}
Jaye~\cite{pluss2023exploring} investigated Q-learning-based market-making strategies in decentralized exchanges (DEXs). The RL agent learns an optimal trading strategy by estimating a Q-function:

\begin{equation}
    Q(s, a) = \mathbb{E} \left[ R(s, a) + \gamma \max_{a'} Q(s', a') \right]
\end{equation}

where:
\begin{itemize}
    \item $s$ is the market state (e.g., liquidity depth, volatility),
    \item $a$ is the action (e.g., placing limit orders, adjusting bid-ask spreads),
    \item $R(s, a)$ is the immediate profit or loss,
    \item $\gamma$ is the discount factor.
\end{itemize}

Findings:
RL-based market-making strategies outperform heuristic-based methods.
The approach was tested in a simulated Uniswap-like environment.
RL learns optimal market-making behavior.

\section{Risk Management in DeFi Lending}

DeFi lending protocols are exposed to various risks, including liquidity risk, collateral risk, oracle risk, and governance risk. Table \ref{tab:defi_risks} outlines the primary risks in DeFi lending.

\begin{table}[ht]
    \centering
    \small % Reduce font size
    \renewcommand{\arraystretch}{1.2} % Adjust row height
    \setlength{\tabcolsep}{5pt} % Adjust column spacing
    \begin{tabular}{|>{\columncolor{gray!20}}l|m{10cm}|}
        \hline
        \textbf{Risk Type} & \textbf{Description} \\
        \hline
        \textbf{Liquidity Risk} & 	If liquidity providers withdraw funds suddenly, borrowers may be unable to repay loans, causing cascading liquidations across multiple protocols.
        \cite{fsb:2023}. \\
        \hline
        \textbf{Collateral Risk} & If collateral prices fall too fast, liquidations may be insufficient to cover outstanding loans, leading to insolvency risks within lending pools. \cite{bis:qt2112b}. \\
        \hline
        \textbf{Oracle Risk} & Manipulated oracles can lead to incorrect liquidation triggers, causing unintended liquidations or price distortions in decentralized exchanges. \cite{gogol2024sokdefi}. \\
        \hline
        \textbf{Flash Loan Risk} & Attackers can exploit flash loans to manipulate markets, conduct arbitrage attacks, or drain liquidity pools without upfront capital. \cite{arxiv:2208.13035}. \\
        \hline
        \textbf{Governance Risk} & Malicious actors can manipulate governance proposals or voting mechanisms, leading to protocol takeovers or fraudulent upgrades.
        \cite{arxiv:2311.17715}. \\
        \hline
    \end{tabular}
    \caption{Types of Risk in DeFi Lending}
    \label{tab:defi_risks}
\end{table}

%% file: sections/98DataPreProcessing.tex
\section{Data Preprocessing}
\label{sec:DataPreprocessing}

The dataset consists of multiple time-series variables, including:

\begin{itemize}
    \item \textbf{Liquidity Metrics:}
    \begin{itemize}
        \item \textit{availableLiquidity}: Amount of liquidity available in the pool.
        \item \textit{totalLiquidity}: Total liquidity supplied to the pool.
        \item \textit{utilizationRate}: Proportion of total liquidity currently borrowed.
    \end{itemize}
    
    \item \textbf{Interest Rate Parameters:}
    \begin{itemize}
        \item \textit{liquidityRate}: Interest rate paid to depositors.
        \item \textit{variableBorrowRate}: Borrow rate for variable-rate loans.
        \item \textit{stableBorrowRate}: Borrow rate for stable-rate loans.
        \item \textit{liquidityIndex}: Cumulative index reflecting liquidity rate over time.
        \item \textit{variableBorrowIndex}: Cumulative index for variable-rate borrowings.
    \end{itemize}

    \item \textbf{Risk Management Variables:}
    \begin{itemize}
        \item \textit{baseLTVasCollateral}: Maximum loan-to-value (LTV) ratio for collateral assets.
        \item \textit{reserveFactor}: Percentage of interest revenue allocated to Aave.
        \item \textit{reserveLiquidationThreshold}: Threshold at which collateral is liquidated.
        \item \textit{isAtRisk}: Binary indicator of whether an asset is at risk of liquidation.
    \end{itemize}
    
    \item \textbf{User Activity Metrics:}
    \begin{itemize}
        \item \textit{lifetimeBorrows}: Total amount borrowed from the pool since inception.
        \item \textit{lifetimeLiquidity}: Total amount deposited into the pool since inception.
        \item \textit{depositVolume}: Daily volume of new deposits.
        \item \textit{borrowVolume}: Daily volume of new borrows.
    \end{itemize}
\end{itemize}

This data is utilized to train an offline reinforcement learning model that optimizes lending parameters by dynamically adjusting interest rates.

\paragraph{Data Preprocessing.}

\begin{enumerate}
    \item \textbf{Timestamp Processing:}  
          The \texttt{\_block\_timestamp} field is converted from UNIX time to a human-readable date format.

    \item \textbf{Normalization of On-Chain Values:}  
          Most values in the dataset are stored as large integers due to Ethereum's decimal precision (10\(^18\)). Each value is scaled down using the appropriate decimal conversion:
          \[
          \text{totalLiquidity} = \frac{\text{totalLiquidity}}{10^{\text{decimals}}}
          \]

    \item \textbf{Computing Interest Rate APY:}  
          Aave's rates are stored as APRs (Annual Percentage Rates). The APY is computed as:
          \[
          APY = \left(1 + \frac{\text{rateAPR}}{365}\right)^{365} - 1
          \]

    \item \textbf{Risk Metric Computation:}
          \begin{itemize}
              \item Loan-to-Value Ratio (LTV) is calculated as:
                \[
                LTV = \frac{\text{Total Debt}}{\text{Total Liquidity as Collateral} + 10^{-6}}
                \]
              \item Liquidation Risk:
                \[
                \text{isAtRisk} = \begin{cases} 
                1, & LTV > \text{reserveLiquidationThreshold} \\
                0, & \text{otherwise} \\
                \end{cases}
                \]
          \end{itemize}

    \item \textbf{Interest Rate Momentum and Volatility:}
          \begin{itemize}
              \item \textit{Momentum} is computed using a rolling average:
                \[
                \text{Momentum}_{t} = \frac{1}{n} \sum_{i=t-n}^{t} R_i
                \]
              \item \textit{Volatility} is calculated as:
                \[
                \text{Volatility}_{t} = \sqrt{\frac{1}{n} \sum_{i=t-n}^{t} (R_i - \bar{R})^2}
                \]
          \end{itemize}
\end{enumerate}

\section{Exploratory Data Analysis (EDA)}
\label{sec:Exploratory} 
\paragraph{Interest Rate Trends}

Understanding the trends in interest rates is crucial for optimizing DeFi lending protocols. The dataset includes time-series information on \textbf{liquidity rates, stable borrow rates, and variable borrow rates} for different assets across Aave V2 and V3. The figures below illustrate these trends.

\begin{figure}[H]
    \centering
    \begin{subfigure}{0.48\textwidth}
        \includegraphics[width=\textwidth]{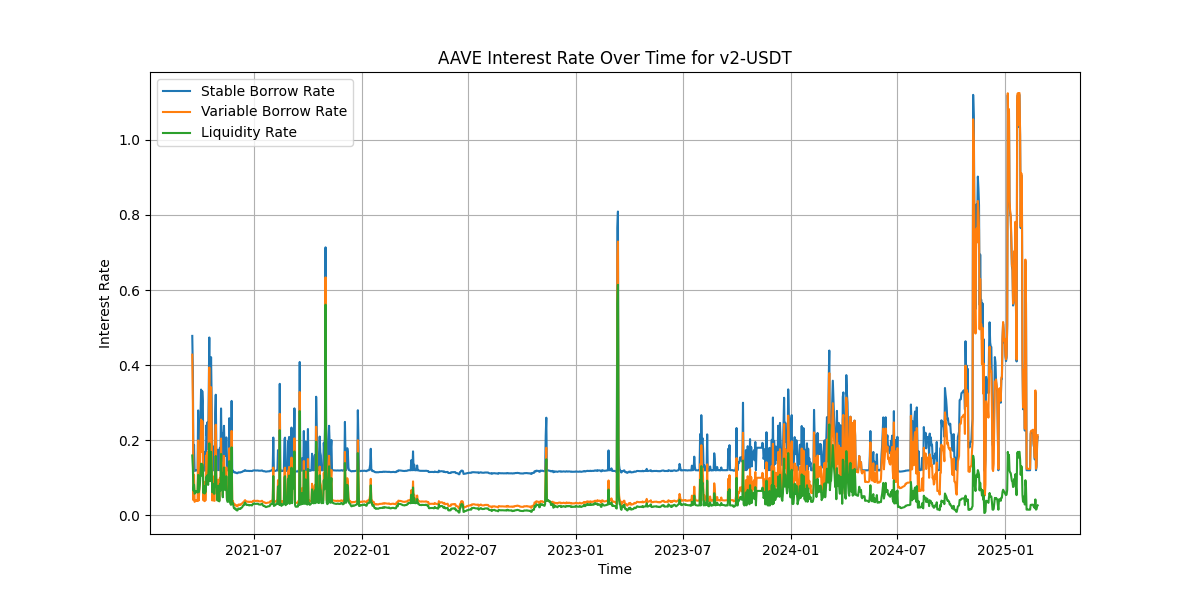}
        \caption{V2 - USDT}
    \end{subfigure}
    \hfill
    \begin{subfigure}{0.48\textwidth}
        \includegraphics[width=\textwidth]{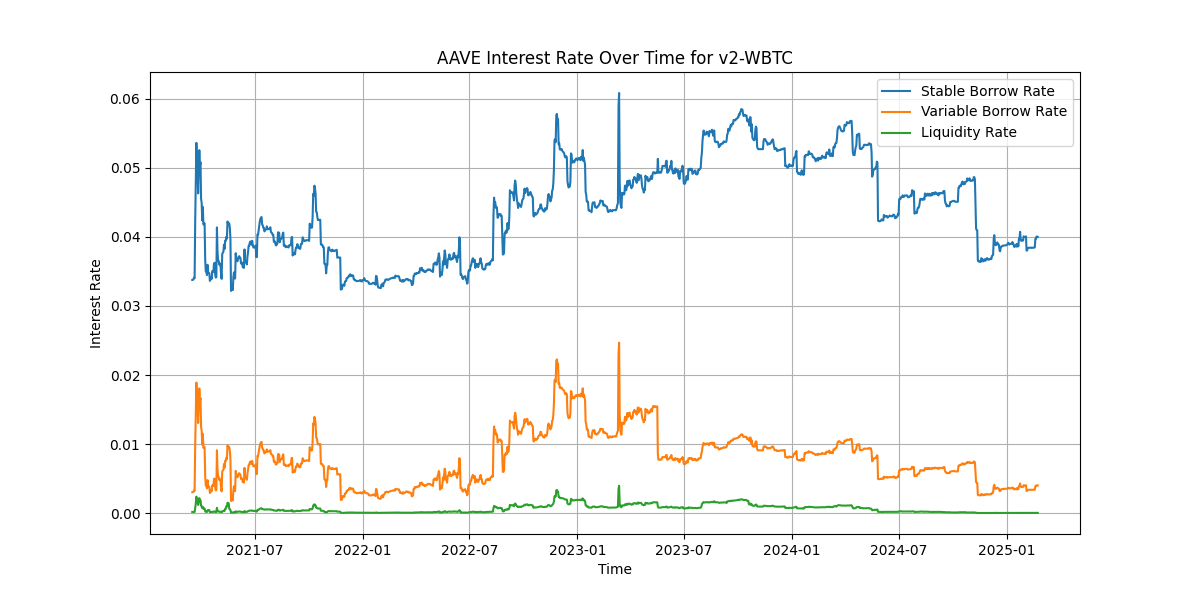}
        \caption{V2 - WBTC}
    \end{subfigure}
    
    \begin{subfigure}{0.48\textwidth}
        \includegraphics[width=\textwidth]{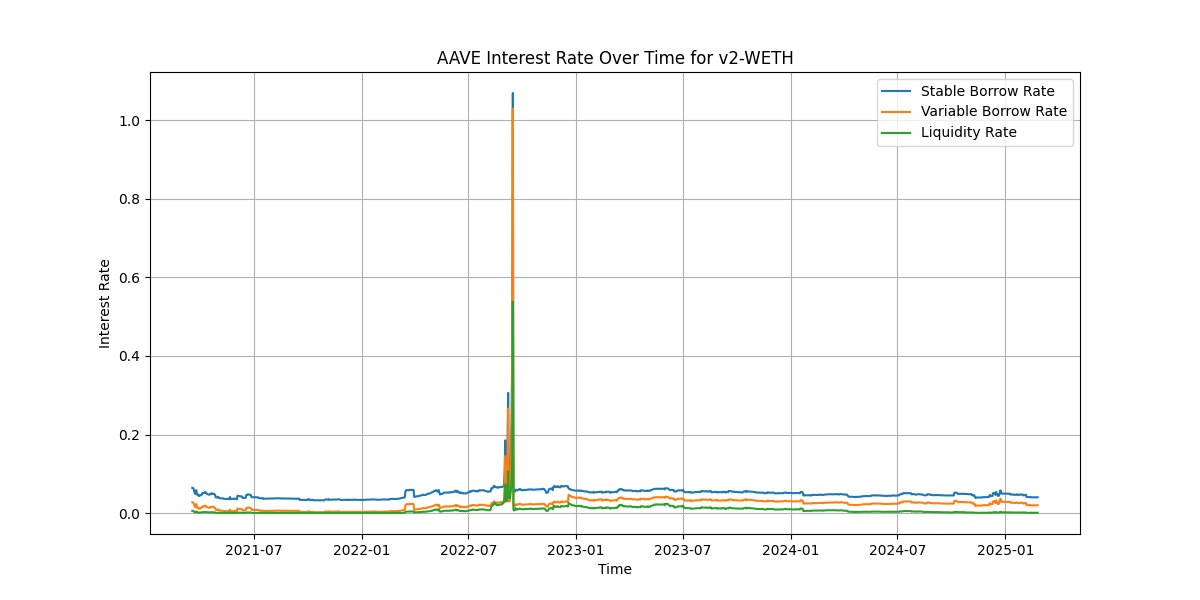}
        \caption{V2 - WETH}
    \end{subfigure}
    \hfill
    \begin{subfigure}{0.48\textwidth}
        \includegraphics[width=\textwidth]{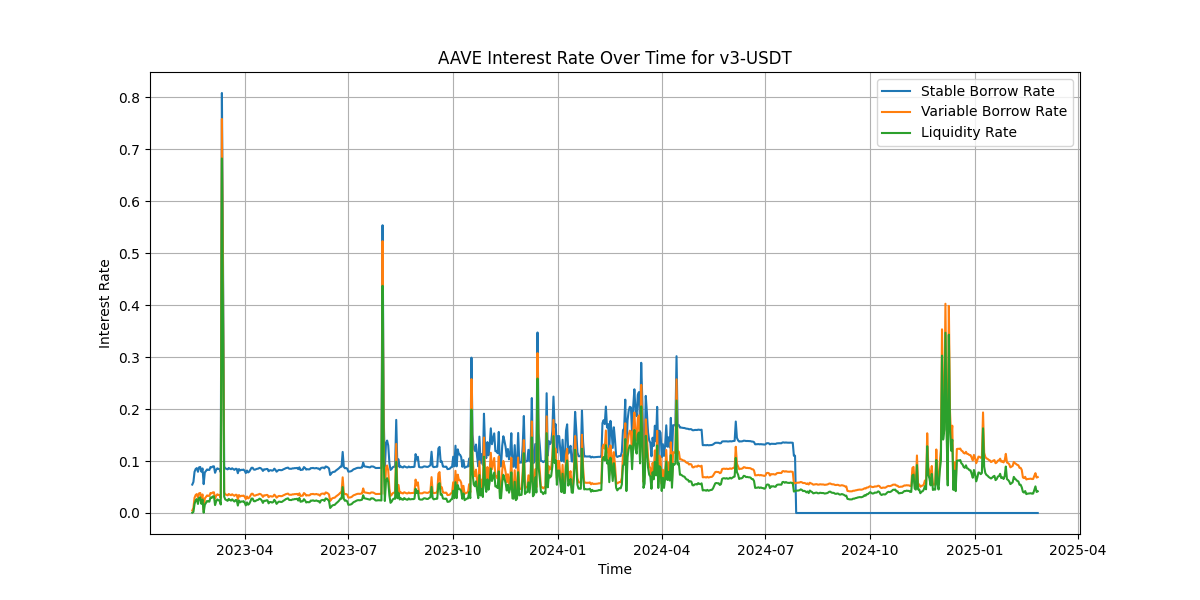}
        \caption{V3 - USDT}
    \end{subfigure}
    
    \begin{subfigure}{0.48\textwidth}
        \includegraphics[width=\textwidth]{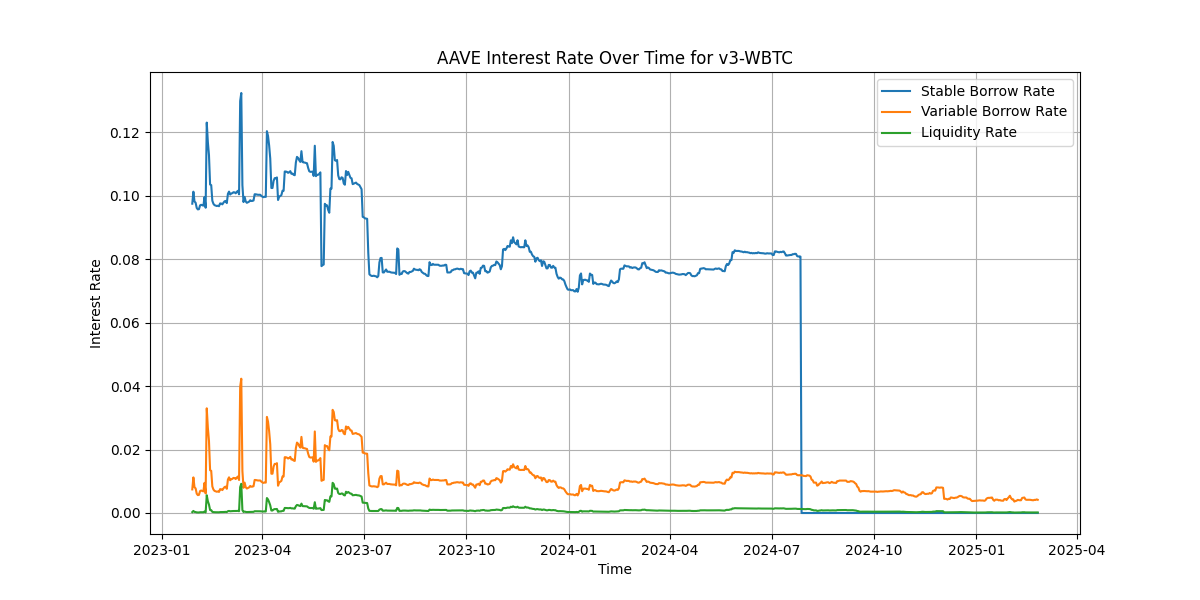}
        \caption{V3 - WBTC}
    \end{subfigure}
    \hfill
    \begin{subfigure}{0.48\textwidth}
        \includegraphics[width=\textwidth]{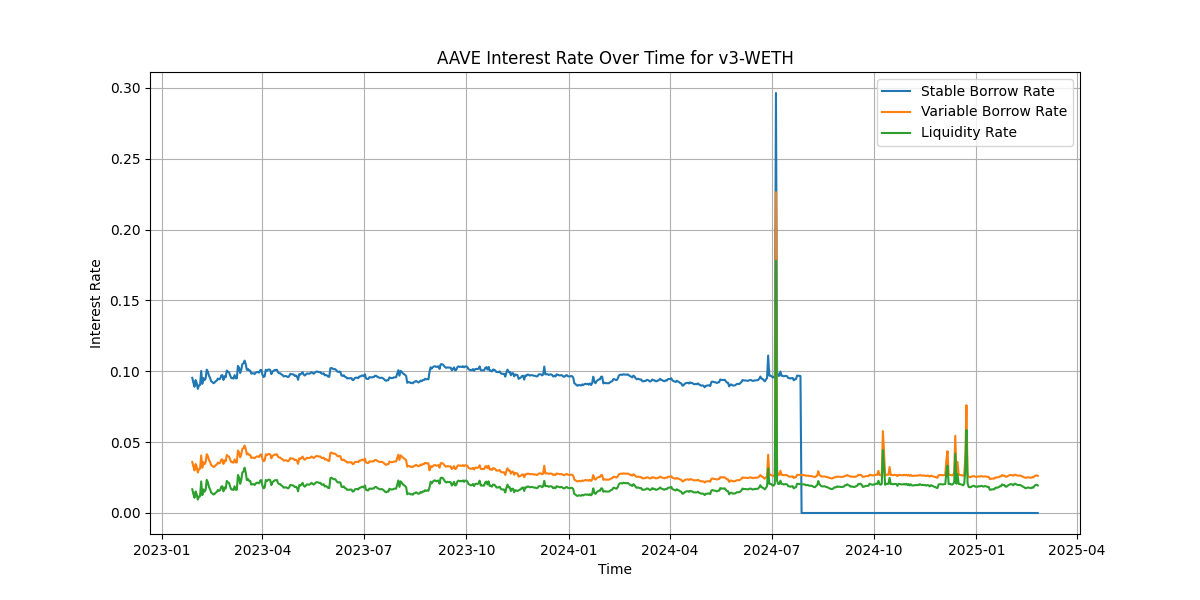}
        \caption{V3 - WETH}
    \end{subfigure}
    
    \caption{Interest Rate Trends Over Time}
    \label{fig:interest_rates}
\end{figure}

The study of interest rate fluctuations across Aave V2 and V3 presents a compelling narrative of market adaptation, liquidity shocks, and strategic governance interventions within decentralized finance (DeFi). Our empirical observations underscore significant variations across different asset classes (USDT, WBTC, and WETH), reflecting changing risk appetites, macroeconomic conditions, and protocol-level optimizations.

\noindent

\textbf{Empirical Observations on Interest Rate Trends}

\textbf{a) Aave V2 Analysis}

\textbf{USDT (Tether)}
\begin{itemize}
    \item Variable Borrow Rate: Characterized by high volatility, with pronounced spikes in late 2024 and early 2025 indicative of liquidity constraints and aggressive borrowing demand surges. These fluctuations correspond to liquidity withdrawals from DeFi, possibly tied to broader market contractions or increased institutional engagement in stablecoin markets.
    \item Stable Borrow Rate: Initially available but later discontinued, aligning with Aave's progressive deprecation of fixed-rate borrowing models in response to liquidity inefficiencies.
    \item Liquidity Rate: Demonstrates a strong correlation with the variable borrow rate, reinforcing the hypothesis that lending yields are driven by borrower-side demand fluctuations rather than consistent deposit inflows.
\end{itemize}

\textbf{WBTC (Wrapped Bitcoin)}
\begin{itemize}
    \item Variable Borrow Rate: Exhibits moderate stability, apart from discrete periods of upward pressure, likely driven by Bitcoin market cycles and arbitrage activities between centralized exchanges and DeFi lending platforms.
    \item Stable Borrow Rate: Phased out in later stages of Aave V2.
    \item Liquidity Rate: Persistently low, suggesting that BTC collateral is predominantly utilized for borrowing leverage rather than yield-generating deposits.
\end{itemize}

\textbf{WETH (Wrapped Ethereum)}
\begin{itemize}
    \item Variable Borrow Rate: Largely stable, except for an anomalous spike in October-November 2022, a period corresponding to the FTX collapse and subsequent liquidity crisis in crypto markets.
    \item Stable Borrow Rate: Discontinued over time, following the same governance-driven rationalization as other assets.
    \item Liquidity Rate: Relatively stable, mirroring the variable borrow rate but with a lagged effect, indicative of market participants responding to yield shifts over time rather than immediately.
\end{itemize}

\textbf{b) Aave V3 Analysis}

\textbf{USDT (Tether)}
\begin{itemize}
    \item Variable Borrow Rate: Experiences a structurally higher baseline compared to V2, with more pronounced fluctuations in late 2024-2025, reflecting evolving capital efficiency mechanisms and potential liquidity constraints.
    \item Liquidity Rate: Closely tracks variable borrowing rates, reinforcing the pro-cyclical nature of liquidity provisioning within Aave's decentralized money markets.
    \item Stable Borrow Rate: No longer supported, marking the full transition to a dynamic, market-driven lending model.
\end{itemize}

\textbf{WBTC (Wrapped Bitcoin)}
\begin{itemize}
    \item Variable Borrow Rate: Shows a steady upward trajectory, potentially linked to growing institutional DeFi participation and strategic portfolio adjustments by long-term BTC holders.
    \item Liquidity Rate: Remains relatively low, suggesting that BTC lending in DeFi continues to be a secondary consideration for most holders compared to centralized finance (CeFi) alternatives.
    \item Stable Borrow Rate: Not included in V3, as per governance-led deprecation efforts.
\end{itemize}

\textbf{WETH (Wrapped Ethereum)}
\begin{itemize}
    \item Variable Borrow Rate: Largely stable, apart from an exceptional spike in October-November 2022, attributed to systemic risk spillovers from the FTX collapse.
    \item Liquidity Rate: Demonstrates delayed responses to extreme borrowing conditions, implying market friction in liquidity adjustments.
    \item Stable Borrow Rate: Eliminated, reinforcing the broader DeFi trend of embracing fully dynamic interest rate mechanisms.
\end{itemize}

\textbf{Systemic Market Events and Governance-Led Adjustments}

\textbf{a) The October-November 2022 Liquidity Shock: Root Causes and Consequences}
A significant short-term interest rate spike in WETH borrowing rates across Aave V2 in October-November 2022 can be attributed to widespread market distress following the collapse of FTX. This collapse had multifaceted repercussions on DeFi liquidity dynamics:

\begin{itemize}
    \item Liquidity Exodus and Flight to Safety: The ensuing market panic led to mass liquidity withdrawals, reducing available lending pools in DeFi platforms.
    \item Heightened Borrowing Demand: Traders and market makers sought liquidity to cover leveraged positions and hedge risk, creating sudden surges in borrowing rates.
    \item Protocol-Level Risk Adjustments: In response, Aave's governance enacted several emergency measures to mitigate systemic risks:
    \begin{itemize}
        \item Interest Rate Curve Adjustments (Nov 30, 2022): Aimed at stabilizing borrowing conditions for assets like USDT and TUSD by modifying rate parameters \cite{aavegov}.
        \item Risk Parameter Modifications (Nov 22, 2022): Gauntlet, Aave's risk management partner, proposed adjustments to collateral and borrowing parameters to mitigate cascading liquidations \cite{unchained}.
        \item Freezing of Low-Liquidity Pools (Nov 28, 2022): Certain illiquid asset pools were frozen to minimize risks and encourage migration to Aave V3 \cite{unchained}.
    \end{itemize}
\end{itemize}

The confluence of user-driven liquidity shocks and governance-led stabilizing actions shaped the observed interest rate fluctuations.

\textbf{b) The Governance Decision to Eliminate Stable Borrowing Rates}

The progressive phasing out of stable borrowing rates across Aave V2 and V3 aligns with both risk-based and efficiency-driven considerations:

\begin{itemize}
    \item Risk Management: Fixed borrowing rates expose the protocol to liquidity mismatch risks, particularly during volatile periods when short-term capital flight can destabilize lending pools.
    \item Capital Efficiency Optimization: Dynamic interest rates allow for real-time liquidity reallocation, leading to more adaptive yield curves that reflect market conditions rather than pre-set constraints.
    \item Governance-Led Rationalization: Proposals such as:
    \begin{itemize}
        \item "BGD: Full Deprecation of Stable Rate Borrowing"
        \item "ARFC: Disable Stable Rate Borrowing for Aave V2 Ethereum Pool"
    \end{itemize}
    These governance votes formalized the strategic shift away from stable borrowing \cite{aavegov}.
\end{itemize}

\textbf{Model Training Implications and Feature Selection Justification}

Given the shift toward fully dynamic borrowing environments, our model training excludes stable borrowing rates and focuses exclusively on:
\begin{itemize}
    \item Variable Borrow Rates: Represent real-time capital demand and cost of liquidity provisioning.
    \item Liquidity Rates: Capture market-driven yield responses and depositor incentives.
\end{itemize}

This selection ensures that our reinforcement learning models align with the contemporary DeFi lending landscape, reflecting the risk-adjusted, dynamic interest rate mechanics that govern borrower and lender behaviors in Aave V3.

\subsubsection{Liquidity and Utilization Analysis}

To assess risk and efficiency, we examine the utilization rate and available liquidity for different reserves. Aave's utilization rate impacts interest rate fluctuations as it determines borrowing incentives.

\begin{equation}
    U = \frac{\text{Total Debt}}{\text{Total Liquidity} + \epsilon}
\end{equation}

where \( U \) is the utilization rate, Total Debt represents the sum of outstanding loans, and Total Liquidity denotes available funds. The term \( \epsilon \) is a small constant to prevent division by zero.

\vspace{10pt}

\subsection{Data Preparation for ML/RL}

\subsubsection{Feature Engineering and Extraction}

The dataset extracted from AaveScan consists of multiple raw features describing the lending protocol's state at different timestamps. We perform feature selection, engineering, and transformation to ensure an efficient learning process for reinforcement learning (RL) models.

\subsubsection*{Selected Features}

The key features used in the dataset are categorized as follows:

\begin{table}[H]
    \centering
    \caption{Feature Engineering for RL Models}
    \label{tab:feature_engineering}
    \begin{tabular}{p{4.5cm} p{8cm}}
        \hline
        \textbf{Category} & \textbf{Features} \\
        \hline
        \textbf{Liquidity Metrics} & Available Liquidity, Total Deposits, Utilization Rate, Total ATokens Supply \\
        \textbf{Debt Metrics} & Total Current Variable Debt, Total Principal Stable Debt, Total Scaled Variable Debt \\
        \textbf{Interest Rates} & Liquidity Rate, Variable Borrow Rate, Stable Borrow Rate, Base Variable Borrow Rate \\
        \textbf{Risk Parameters} & Loan-To-Value (LTV), Reserve Factor, Liquidation Threshold, Reserve Liquidation Bonus \\
        \textbf{Market Indicators} & Price of WBTC/WETH, Deposit-to-Borrow Ratio, Liquidity Volatility \\
        \textbf{Historical Trends} & 7-day rolling average of liquidity rates, 7-day volatility of borrow rates \\
        \hline
    \end{tabular}
\end{table}

These features serve as the basis for state representation in RL-based interest rate optimization.

\subsubsection*{Feature Transformations}

Several transformations are applied to improve model performance:

\begin{itemize}
    \item \textbf{Normalization}: Features with different scales are normalized for better learning efficiency.
    \item \textbf{Rolling Mean Computation}: Time-series features (interest rates, borrow volume) are smoothed using a 7-day moving average.
    \item \textbf{Log Transformation}: Applied to highly skewed variables such as lifetime liquidations and flash loan volumes.
    \item \textbf{Utilization Rate Adjustment}: Defined as:

    \begin{equation}
        U_t = \frac{\text{Total Debt}_t}{\text{Total Liquidity}_t + \epsilon}
    \end{equation}

    where \( U_t \) is the utilization rate at time \( t \), and \( \epsilon \) is a small constant to avoid division by zero.
\end{itemize}

%% file: sections/99Appendix.tex
\section{Training Performance Evaluation: CQL, BC, and TD3BC}
\label{sec:PerformanceEvluation}

\subsection*{Understanding Actor and Critic Losses}
Reinforcement learning (RL) models are typically evaluated based on their ability to learn an optimal policy while maintaining stable training dynamics. In this section, we analyze the training performance of three different RL-based approaches for optimizing DeFi lending parameters: Conservative Q-Learning (CQL), Behavior Cloning (BC), and Twin Delayed Deep Deterministic Policy Gradient with Behavior Cloning (TD3-BC).

A key aspect of evaluating RL models is analyzing the \textbf{actor loss} and \textbf{critic loss}, which provide insights into how well the model is optimizing its policy and estimating value functions. Understanding these losses allows us to diagnose training stability and policy efficiency~\cite{fujimoto2021minimalist,torabi2018bc}.

\subsubsection*{Actor Loss: Policy Optimization and Convergence}

The \textbf{actor loss} measures how effectively the policy (actor) learns to select actions that maximize long-term rewards. It is typically defined as:

\begin{equation}
L_{\text{actor}} = -\mathbb{E}_{s_t \sim D} \left[ Q(s_t, \pi_\theta(s_t)) \right]
\end{equation}

where \( Q(s_t, a) \) is the critic's estimate of the expected reward for taking action \( a \) in state \( s_t \). Since the actor optimizes for high Q-values, a lower actor loss implies a better policy.

A good actor loss should:
\begin{itemize}
    \item Decrease over time and stabilize near zero, indicating policy convergence.
    \item Avoid excessive fluctuations, as instability may indicate weak critic guidance or poor reward scaling~\cite{fujimoto2021minimalist}.
    \item Not drop too quickly, as rapid convergence suggests overfitting to historical data rather than learning an optimal policy.
\end{itemize}

\textbf{Actor Loss Behavior in Different Models:}
\begin{itemize}
    \item \textbf{BC}: The actor loss drops immediately to near zero since it merely mimics historical actions~\cite{torabi2018bc}.
    \item \textbf{CQL}: Actor loss may increase initially due to restrictive Q-learning constraints but should eventually stabilize~\cite{kumar2020cql}.
    \item \textbf{TD3-BC}: Actor loss should steadily decrease and stabilize as the model finds a balance between imitation and reinforcement learning~\cite{fujimoto2021minimalist}.
\end{itemize}

A well-trained RL model should have a smoothly decreasing actor loss that stabilizes near zero without excessive fluctuations.

\subsubsection*{Critic Loss: Value Function Stability}

The \textbf{critic loss} measures how accurately the Q-values approximate expected rewards. It is defined as:

\begin{equation}
L_{\text{critic}} = \mathbb{E}_{(s,a,r,s') \sim D} \left[ (Q(s, a) - y)^2 \right]
\end{equation}

where:

\begin{equation}
y = r + \gamma \min_{i=1,2} Q(s', \pi(s'))
\end{equation}

A good critic loss should:
\begin{itemize}
    \item Decrease steadily and stabilize, ensuring consistent Q-value estimation.
    \item Not diverge, as an increasing loss suggests overestimation of Q-values.
    \item Not oscillate excessively, as high fluctuations indicate an unstable value function~\cite{fujimoto2018td3}.
\end{itemize}

\textbf{Critic Loss Behavior in Different Models:}
\begin{itemize}
    \item \textbf{BC}: No critic loss since there is no Q-learning component.
    \item \textbf{CQL}: Critic loss initially increases due to conservative penalties but later stabilizes~\cite{kumar2020cql}.
    \item \textbf{TD3-BC}: Critic loss steadily decreases and remains stable, ensuring reliable Q-value estimation~\cite{fujimoto2021minimalist}.
\end{itemize}

A well-trained RL model should have a critic loss that steadily decreases and stabilizes, ensuring accurate value estimation.

\subsection{Conservative Q-Learning (CQL)}

CQL aims to prevent overestimation in Q-learning by enforcing conservative Q-value updates. In training, the critic loss is expected to increase significantly before stabilizing.

\textbf{Critic loss evaluation}
\begin{figure}[H]
    \centering
    \begin{subfigure}{0.48\textwidth}
        \centering
        \includegraphics[width=\textwidth]{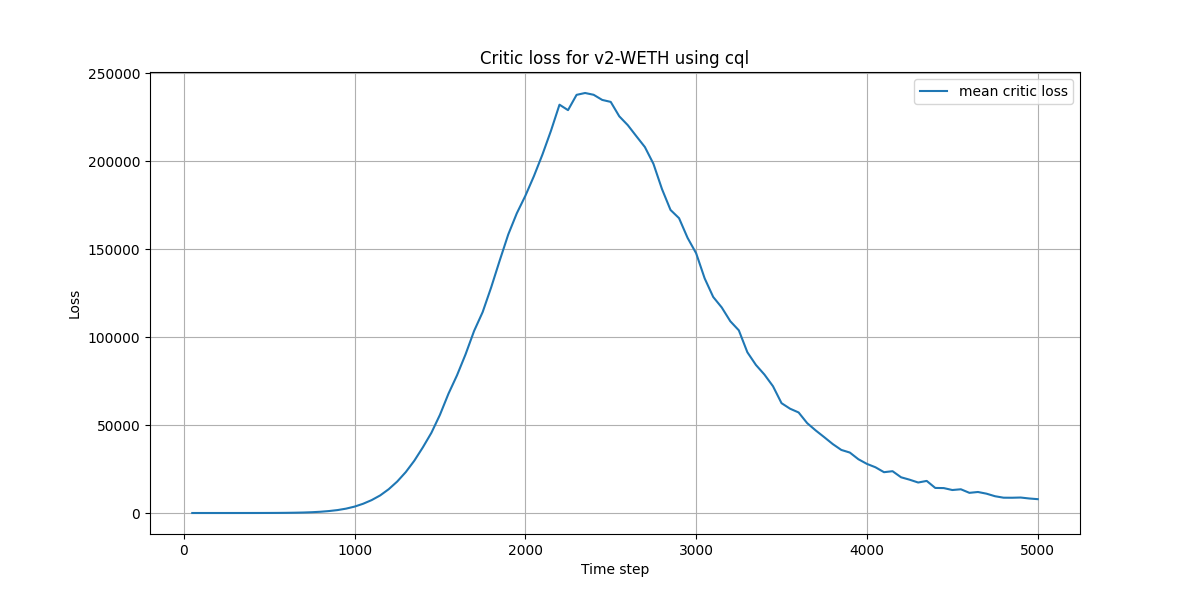}
        \caption{V2 WETH CQL Critic Loss}
    \end{subfigure}
    \begin{subfigure}{0.48\textwidth}
        \centering
        \includegraphics[width=\textwidth]{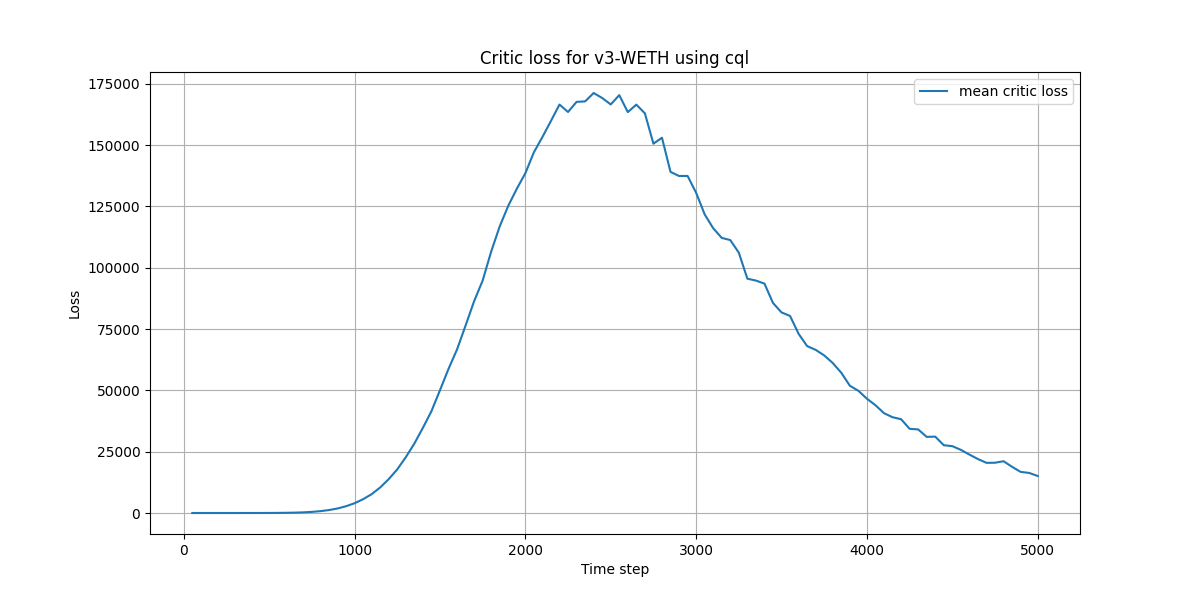}
        \caption{V3 WETH CQL Critic Loss}
    \end{subfigure}
    \caption{Comparison of critic loss for WETH using CQL}
\end{figure}

\begin{figure}[H]
    \centering
    \begin{subfigure}{0.48\textwidth}
        \centering
        \includegraphics[width=\textwidth]{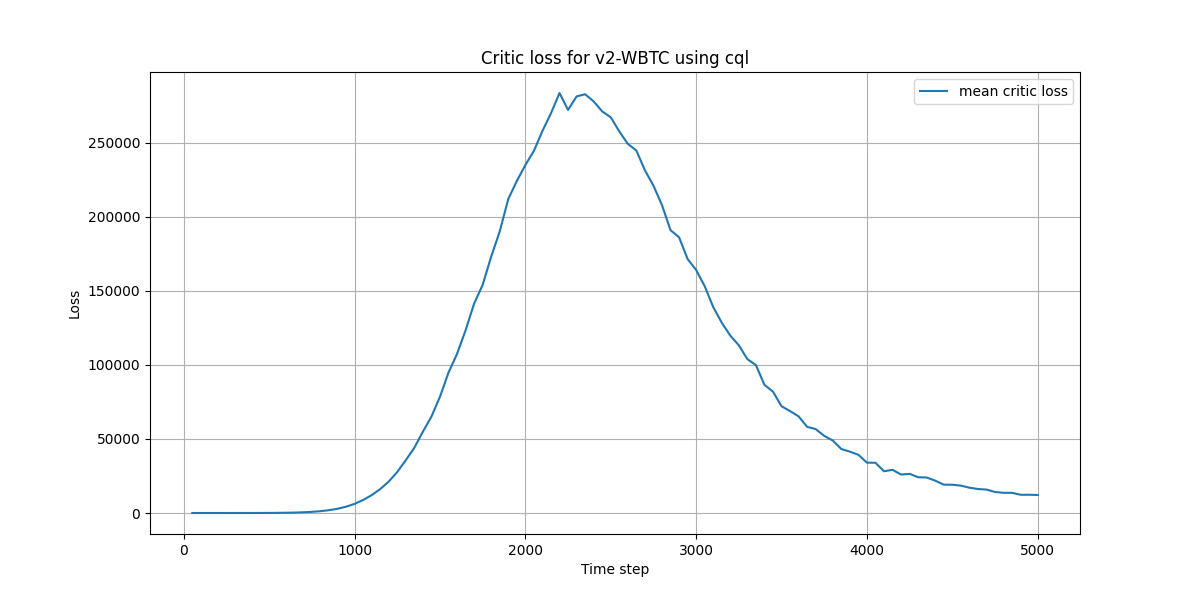}
        \caption{V2 WBTC CQL Critic Loss}
    \end{subfigure}
    \begin{subfigure}{0.48\textwidth}
        \centering
        \includegraphics[width=\textwidth]{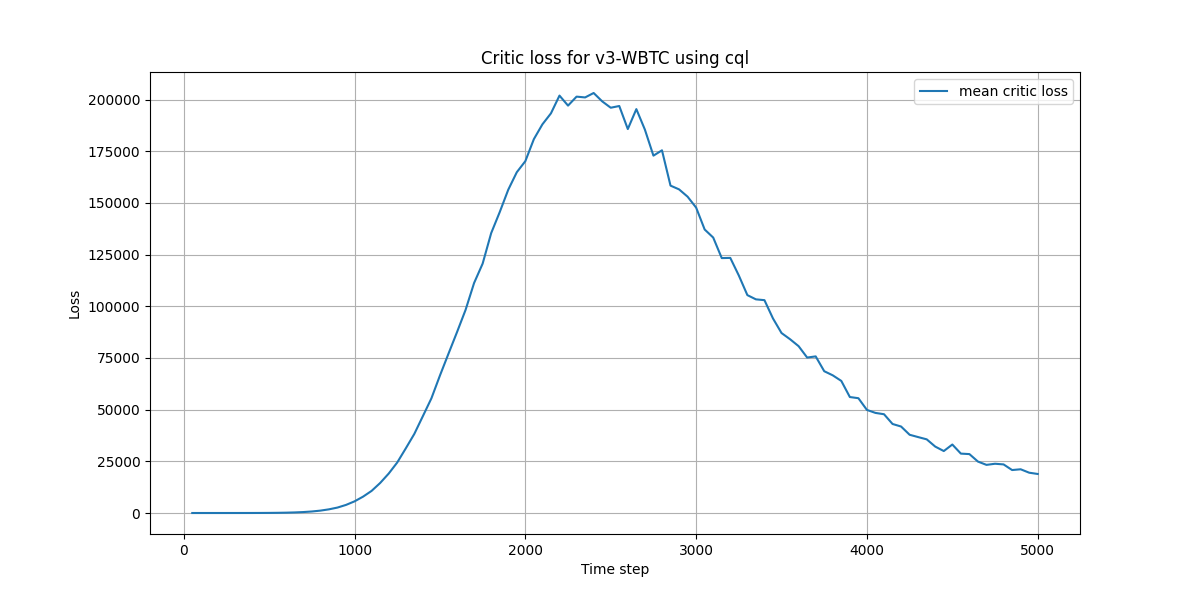}
        \caption{V3 WBTC CQL Critic Loss}
    \end{subfigure}
    \caption{Comparison of critic loss for WBTC using CQL}
\end{figure}

For both Aave V2 and V3, the critic loss exhibits a sharp increase at the beginning of training, reaches a peak, and then gradually declines. This pattern suggests that the Q-function starts with relatively small values, then expands aggressively as the model tries to approximate the true value function, before eventually stabilizing.
\begin{itemize}
    \item WETH: In V2, the critic loss follows a smoother trajectory, indicating that the model is able to approximate Q-values with relatively less fluctuation compared to V3.In V3, the peak critic loss is significantly higher, suggesting that the model struggles more with adjusting to the loan dynamics in V3, possibly due to different liquidity patterns or borrowing demand variations.
    \item WBTC: A similar pattern is observed for WBTC, but the critic loss in V3 has a higher magnitude and takes longer to stabilize, reflecting increased difficulty in learning optimal Q-values. This may indicate that BTC-backed loans involve more complex lending behaviors, possibly due to higher price volatility compared to ETH-backed loans.
\end{itemize}

A key takeaway from the critic loss behavior is that CQL struggles with convergence in both V2 and V3, with V3 being more unstable. This is consistent with CQL's known issues in settings where the reward landscape is highly dynamic, as is often the case in DeFi lending.

\textbf{Actor loss evaluation}

Unlike TD3BC or BC, where the actor loss decreases steadily, CQL actor loss increases over time before stabilizing at a high value. This suggests that the policy network struggles to find optimal actions under the conservative Q-function constraints.

\begin{figure}[H]
    \centering
    \begin{subfigure}{0.48\textwidth}
        \centering
        \includegraphics[width=\textwidth]{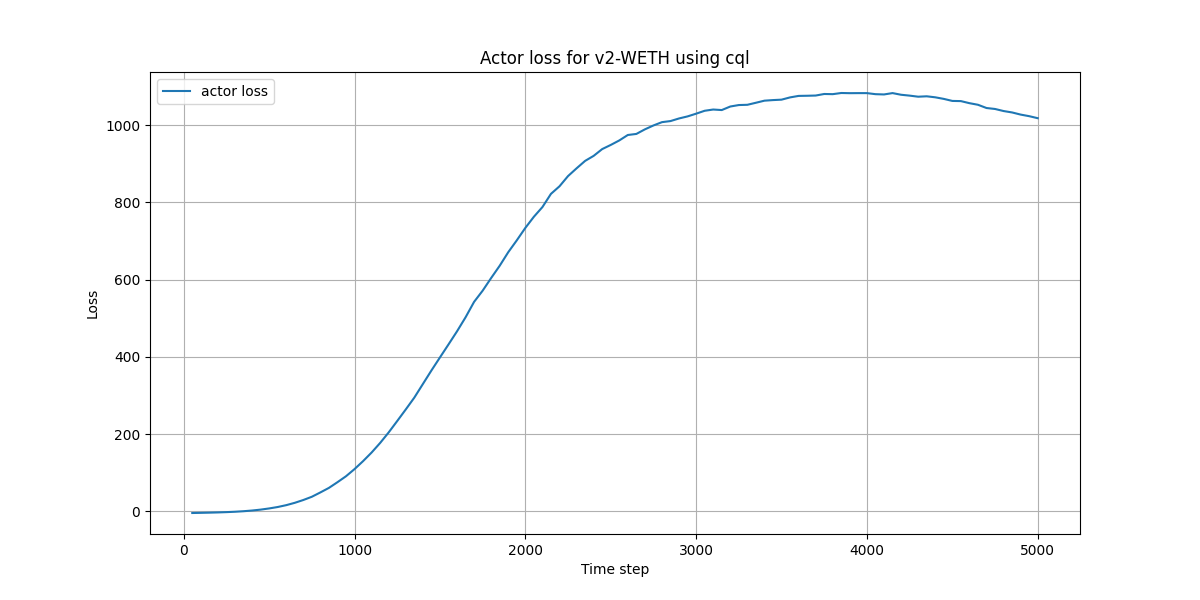}
        \caption{V2 WETH CQL Actor Loss}
    \end{subfigure}
    \begin{subfigure}{0.48\textwidth}
        \centering
        \includegraphics[width=\textwidth]{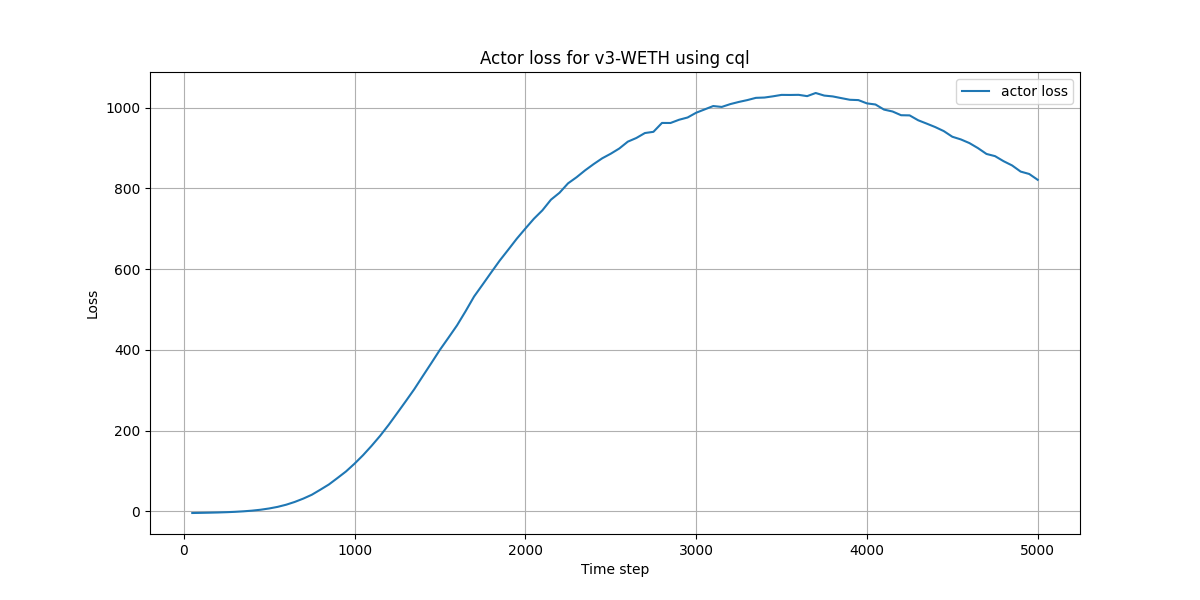}
        \caption{V3 WETH CQL Actor Loss}
    \end{subfigure}
    \caption{Comparison of actor loss for WETH using CQL}
\end{figure}

\begin{figure}[H]
    \centering
    \begin{subfigure}{0.48\textwidth}
        \centering
        \includegraphics[width=\textwidth]{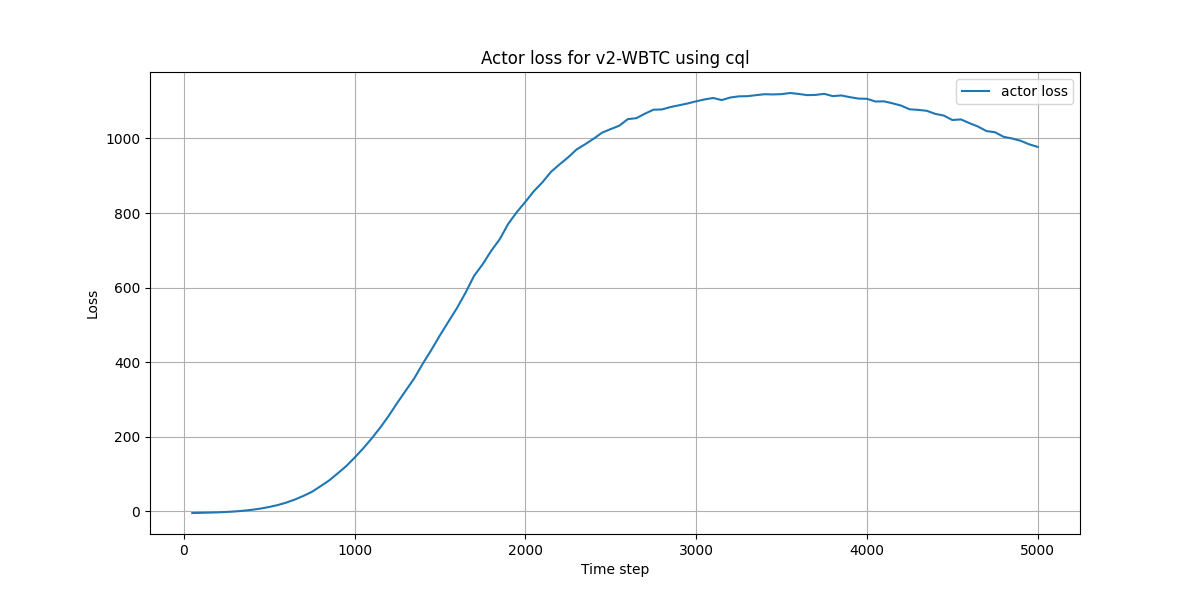}
        \caption{V2 WBTC CQL Actor Loss}
    \end{subfigure}
    \begin{subfigure}{0.48\textwidth}
        \centering
        \includegraphics[width=\textwidth]{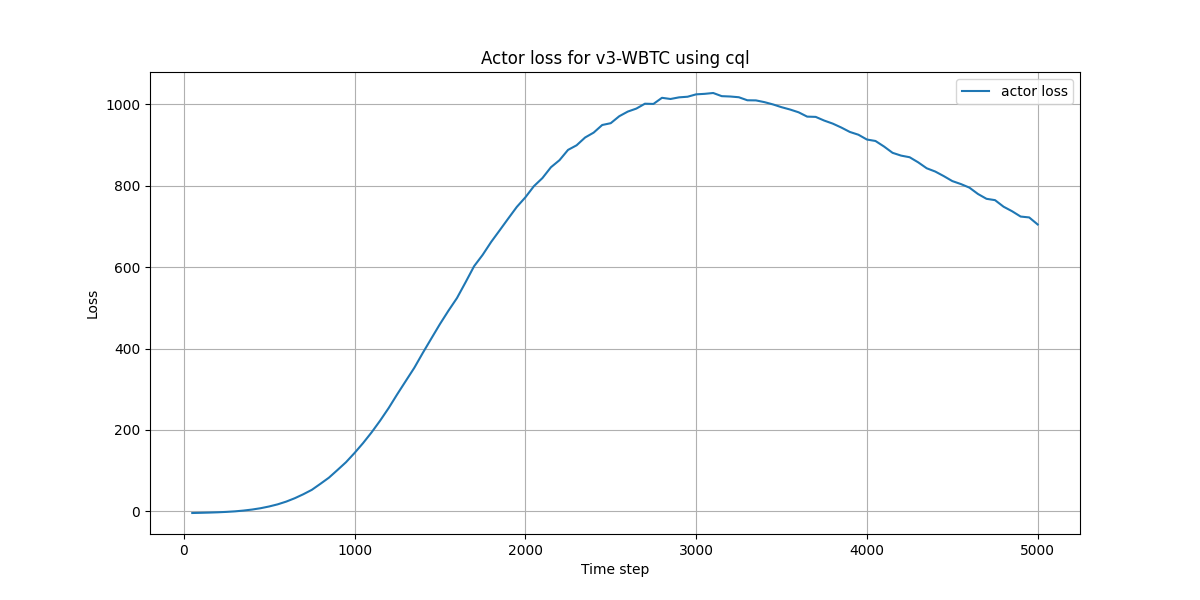}
        \caption{V3 WBTC CQL Actor Loss}
    \end{subfigure}
    \caption{Comparison of actor loss for WBTC using CQL}
\end{figure}

Unlike TD3BC or BC, where the actor loss decreases steadily, CQL actor loss increases over time before stabilizing at a high value. This suggests that the policy network struggles to find optimal actions under the conservative Q-function constraints.
\begin{itemize}
    \item In WETH and WBTC, V2 shows a more controlled increase in actor loss compared to V3, further confirming that V3's environment is more challenging for learning stable policies.
    \item The slow stabilization of actor loss indicates that CQL sacrifices policy efficiency for risk control, which could be beneficial in high-risk lending environments but may lead to overly restrictive policies that limit capital efficiency.
\end{itemize}

\textbf{CQL Summary}
\begin{itemize}
    \item Strengths: Provides a conservative lending strategy, which may reduce exposure to bad debt risks. Can prevent excessive borrowing incentives caused by overoptimistic Q-values.
    \item Weaknesses: Unstable critic loss and slow policy adaptation limit performance. May be too restrictive in some scenarios, reducing capital efficiency.
\end{itemize}

\subsection{Behavioral Cloning (BC)}

\begin{figure}[H]
    \centering
    \begin{subfigure}{0.48\textwidth}
        \centering
        \includegraphics[width=\textwidth]{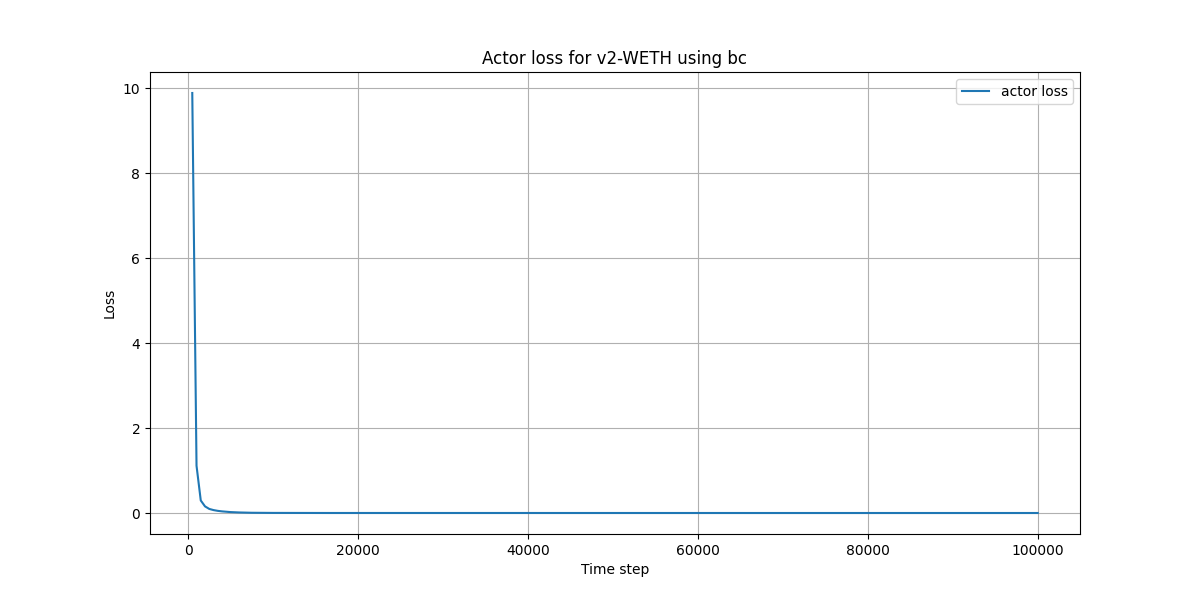}
        \caption{V2 WETH BC Actor Loss}
    \end{subfigure}
    \begin{subfigure}{0.48\textwidth}
        \centering
        \includegraphics[width=\textwidth]{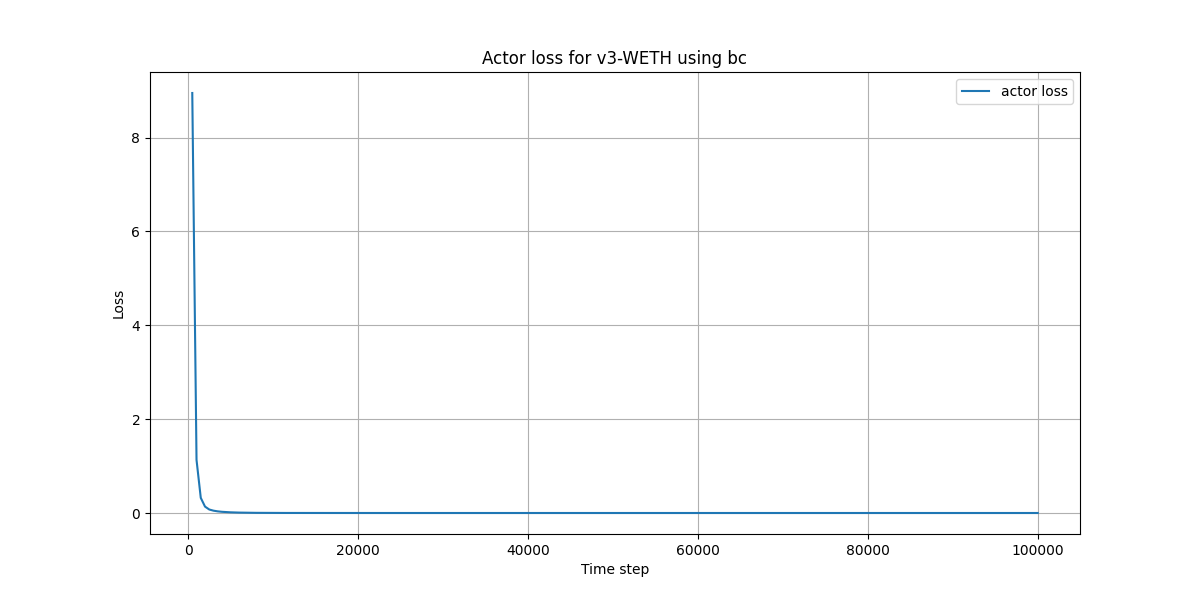}
        \caption{V3 WETH BC Actor Loss}
    \end{subfigure}
    \caption{Comparison of actor loss for WETH using BC}
\end{figure}

\begin{figure}[H]
    \centering
    \begin{subfigure}{0.48\textwidth}
        \centering
        \includegraphics[width=\textwidth]{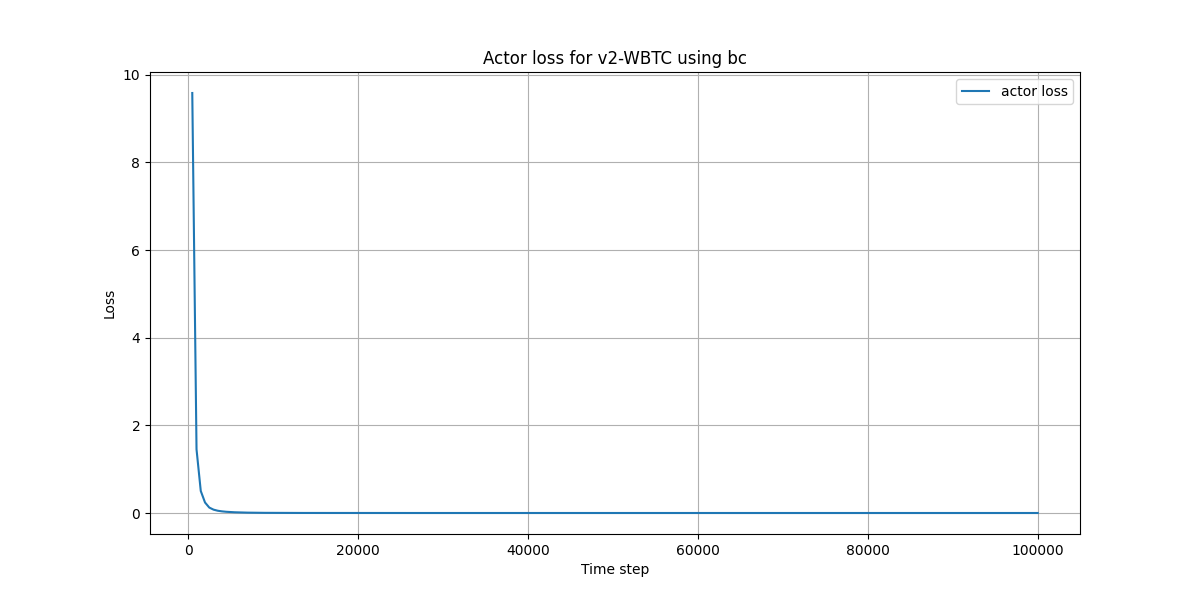}
        \caption{V2 WBTC BC Actor Loss}
    \end{subfigure}
    \begin{subfigure}{0.48\textwidth}
        \centering
        \includegraphics[width=\textwidth]{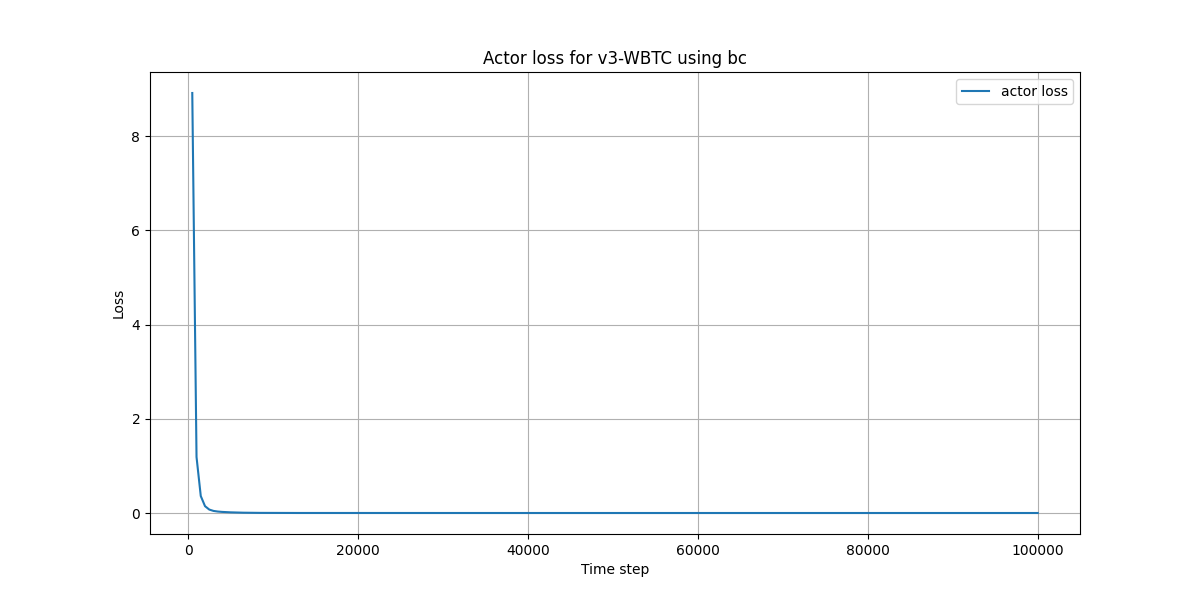}
        \caption{V3 WBTC BC Actor Loss}
    \end{subfigure}
    \caption{Comparison of actor loss for WBTC using BC}
\end{figure}

BC provides a non-reinforcement-learning baseline by directly imitating historical lending behaviors. Unlike CQL or TD3BC, BC does not optimize for long-term performance, making it a useful reference point but not a standalone solution.

\textbf{Actor Loss Evaluation}

BC actor loss drops to near-zero almost immediately, which is expected because the model is directly mimicking observed actions rather than optimizing them.
\begin{itemize}
    \item In both V2 and V3, the loss converges to zero quickly for WETH and WBTC, indicating that BC is able to learn a policy that closely resembles past lending behaviors.
    \item However, this also means BC cannot adapt to new market conditions, making it a poor choice for optimizing lending performance in evolving DeFi environments.
\end{itemize}

\textbf{Limitations of BC}

While BC is useful for benchmarking RL-based models, its major limitation is its inability to generalize beyond the training data.

If past lending policies were inefficient or suboptimal, BC will simply reproduce those inefficiencies.

BC does not account for changes in interest rate mechanisms, liquidity shifts, or evolving borrower behavior, making it an inflexible solution in DeFi.

\textbf{BC Summary}
\begin{itemize}
    \item Strengths:
Simple, fast, and requires minimal computation.
Provides a baseline for evaluating reinforcement learning models.
    \item Weaknesses:
Does not optimize lending policies beyond historical behavior.
Fails to adapt to new market conditions, making it unsuitable for real-world deployment.
\end{itemize}

\subsection{Twin Delayed Deep Deterministic Policy Gradient with Behavior Cloning (TD3-BC)}

TD3-BC combines behavioral cloning with reinforcement learning to strike a balance between policy imitation and strategic optimization. This approach allows the model to leverage historical data while still optimizing for better lending decisions.

\textbf{Critic Loss Evaluation}
\begin{figure}[H]
    \centering
    \begin{subfigure}{0.48\textwidth}
        \centering
        \includegraphics[width=\textwidth]{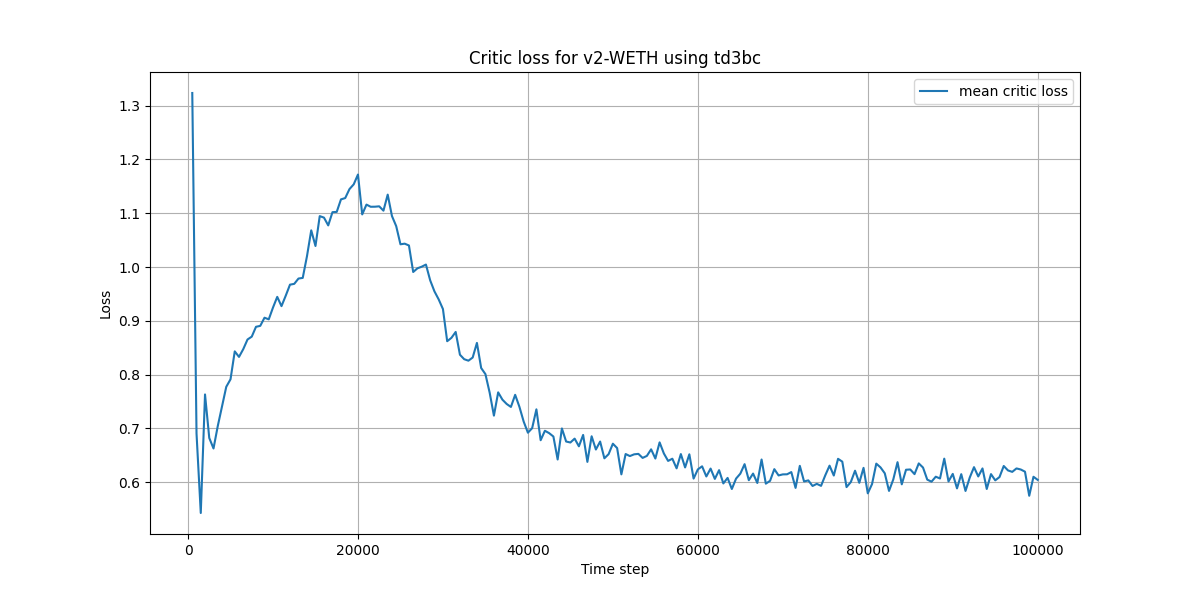}
        \caption{V2 WETH TD3BC Critic Loss}
    \end{subfigure}
    \begin{subfigure}{0.48\textwidth}
        \centering
        \includegraphics[width=\textwidth]{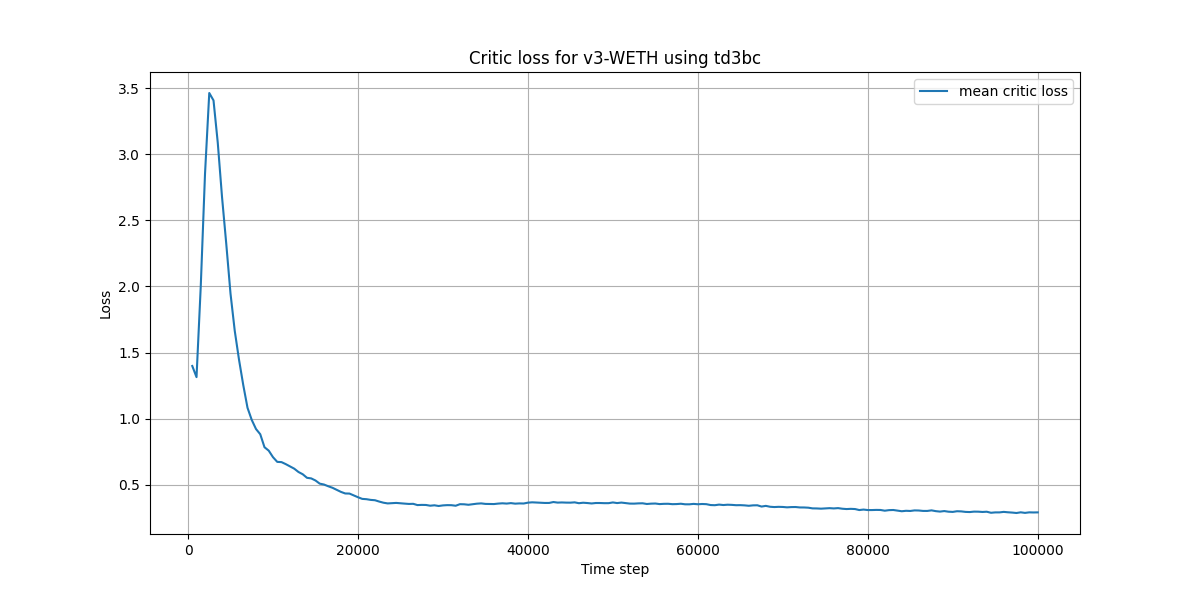}
        \caption{V3 WETH TD3BC Critic Loss}
    \end{subfigure}
    \caption{Comparison of critic loss for WETH using TD3BC}
\end{figure}

\begin{figure}[H]
    \centering
    \begin{subfigure}{0.48\textwidth}
        \centering
        \includegraphics[width=\textwidth]{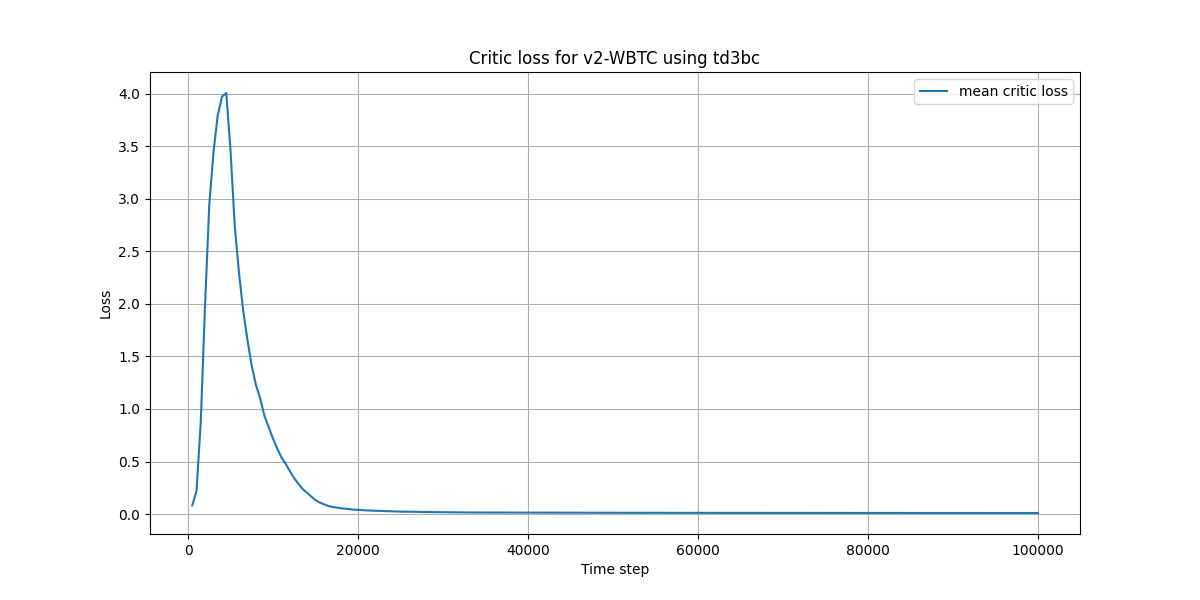}
        \caption{V2 WBTC TD3BC Critic Loss}
    \end{subfigure}
    \begin{subfigure}{0.48\textwidth}
        \centering
        \includegraphics[width=\textwidth]{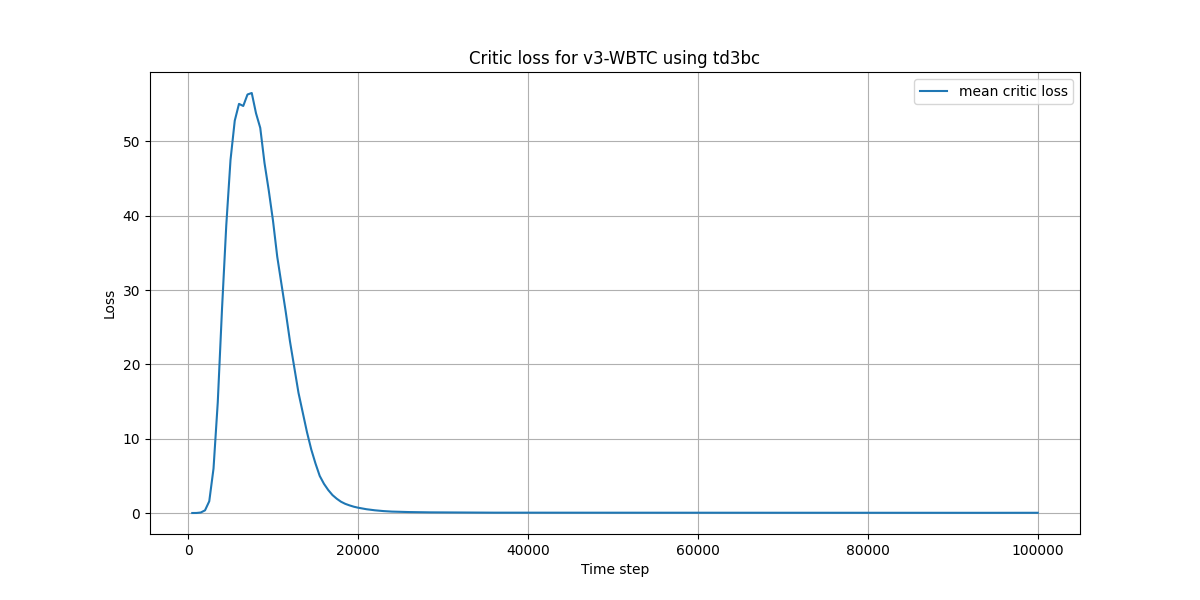}
        \caption{V3 WBTC TD3BC Critic Loss}
    \end{subfigure}
    \caption{Comparison of critic loss for WBTC using TD3BC}
\end{figure}

Unlike CQL, TD3BC critic loss decreases smoothly and stabilizes quickly, demonstrating stable Q-value learning.

For WETH and WBTC, both V2 and V3 show a steady decline in critic loss, with no significant spikes. The smoothness of the curve suggests that TD3BC effectively learns accurate Q-values while maintaining training stability. V3 critic loss is slightly higher than V2, but the difference is much smaller compared to CQL, indicating better adaptability to V3 dynamics.

\textbf{Actor Loss Evaluation}

TD3-BC actor loss drops rapidly and stabilizes near-zero, similar to BC, but with a key difference:

Unlike BC, TD3-BC optimizes the policy while learning, rather than just mimicking past actions.The rapid drop in actor loss indicates that the policy quickly finds optimal lending strategies while incorporating reinforcement learning improvements. V3 actor loss remains slightly higher than V2, suggesting that TD3-BC still faces some challenges in optimizing lending decisions in the more complex V3 environment.

\begin{figure}[H]
    \centering
    \begin{subfigure}{0.48\textwidth}
        \centering
        \includegraphics[width=\textwidth]{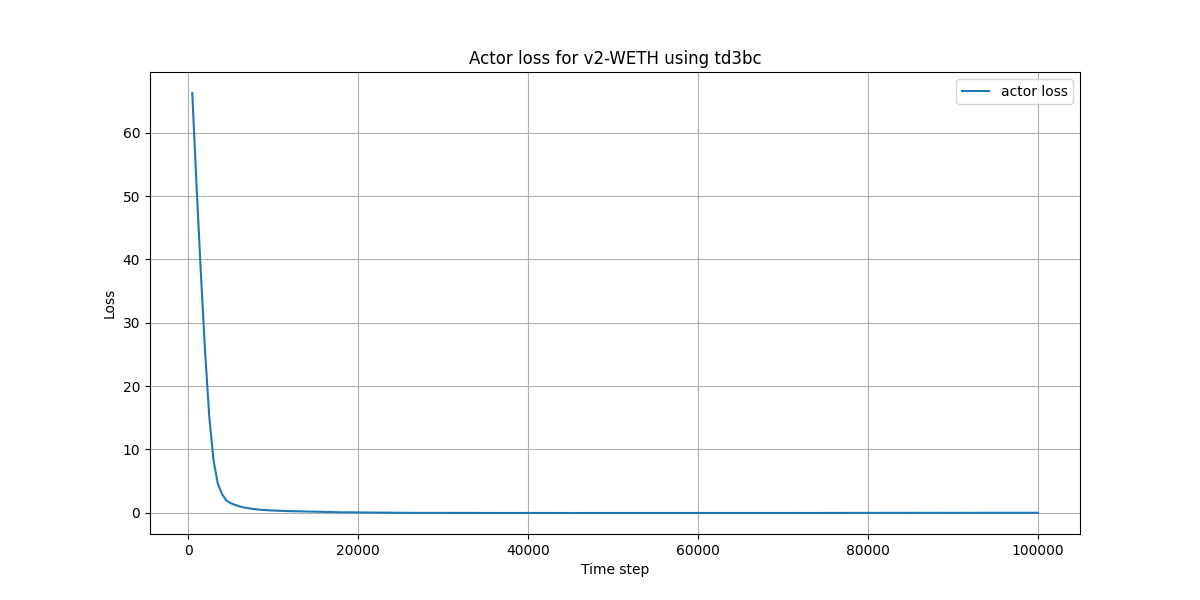}
        \caption{V2 WETH TD3BC Actor Loss}
    \end{subfigure}
    \begin{subfigure}{0.48\textwidth}
        \centering
        \includegraphics[width=\textwidth]{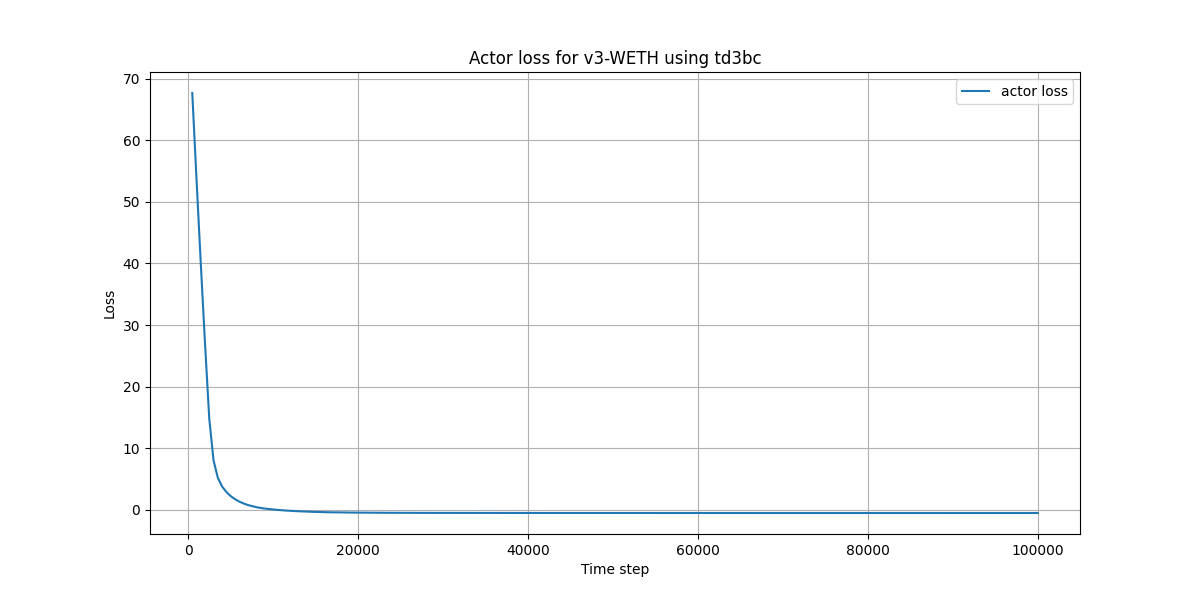}
        \caption{V3 WETH TD3BC Actor Loss}
    \end{subfigure}
    \caption{Comparison of actor loss for WETH using TD3BC}
\end{figure}

\begin{figure}[H]
    \centering
    \begin{subfigure}{0.48\textwidth}
        \centering
        \includegraphics[width=\textwidth]{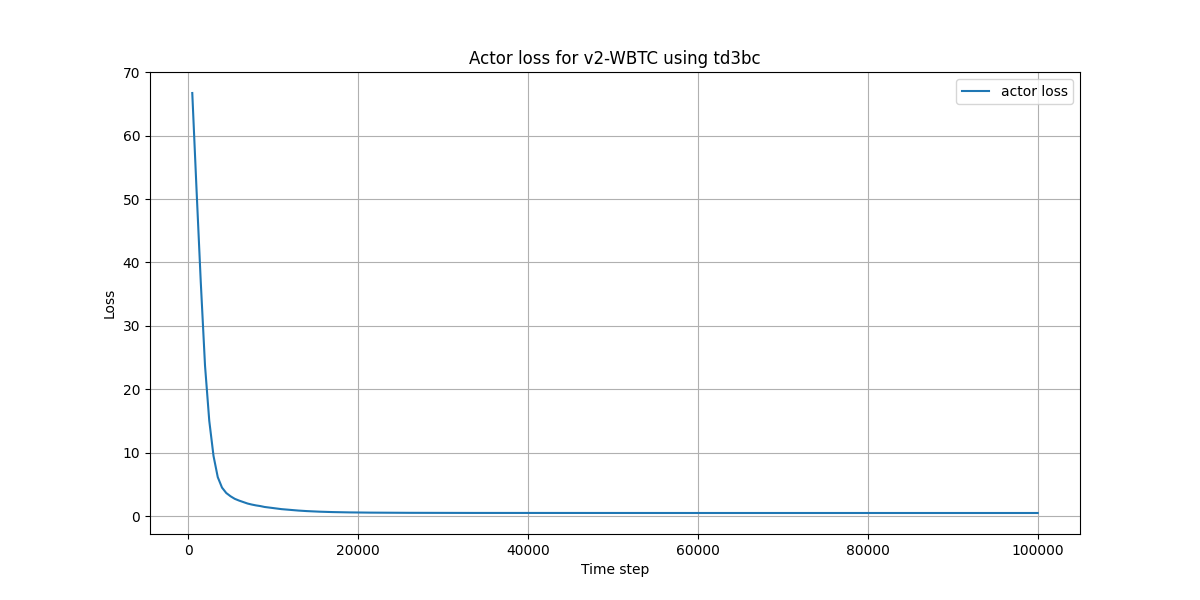}
        \caption{V2 WBTC TD3BC Actor Loss}
    \end{subfigure}
    \begin{subfigure}{0.48\textwidth}
        \centering
        \includegraphics[width=\textwidth]{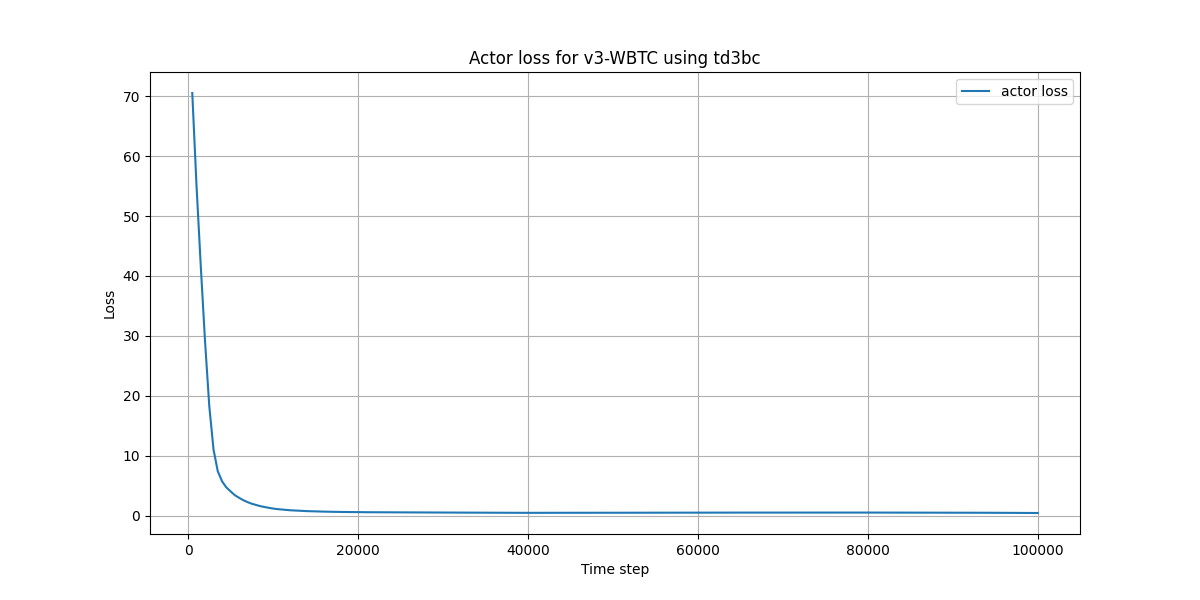}
        \caption{V3 WBTC TD3BC Actor Loss}
    \end{subfigure}
    \caption{Comparison of actor loss for WBTC using TD3BC}
\end{figure}

\textbf{TD3-BC Advantages}
\begin{itemize}
    \item Unlike CQL, TD3BC does not suffer from unstable Q-value updates, making it a more practical choice for real-world deployment.
    \item Unlike BC, TD3BC does not just replicate past behaviors-it actively optimizes lending policies for better capital efficiency and profitability.
    \item Balances exploration and exploitation, allowing for better generalization across different lending environments.
\end{itemize}

\textbf{TD3-BC Summary}
\begin{itemize}
    \item Strengths:
Stable Q-learning with smooth convergence.
Optimized balance of behavior imitation and RL-based improvement.
Generalizes well across Aave V2 and V3.
    \item Weaknesses:
Still influenced by the quality of historical data (if past lending policies were inefficient, TD3BC may inherit some of those inefficiencies).
Slightly higher actor loss in V3 suggests room for improvement in adapting to more complex market conditions.
\end{itemize}

\subsection{Interest Rate Volatility}

\begin{table}[H]
\centering
\caption{Interest Rate Volatility Comparison (Aave vs TD3-BC)}
\label{tab:volatility_comparison}
\begin{tabular}{lllll}
\toprule
\textbf{Asset-Protocol} & \textbf{Rate Type} & \textbf{Aave Std. Dev.} & \textbf{TD3-BC Std. Dev.} & \textbf{Increasing} \\
\midrule
V2-WETH & Borrow Rate     & 3.31E-02 & 3.33E-02 & 0.73\% \\
V2-WETH & Liquidity Rate  & 1.77E-02 & 1.78E-02 & 0.45\% \\
V2-WBTC & Borrow Rate     & 7.93E-04 & 7.97E-04 & 0.59\% \\
V2-WBTC & Liquidity Rate  & 1.40E-04 & 1.41E-04 & 0.54\% \\
V3-WETH & Borrow Rate     & 2.32E-03 & 2.36E-03 & 1.57\% \\
V3-WETH & Liquidity Rate  & 8.52E-03 & 8.58E-03 & 0.71\% \\
V3-WBTC & Borrow Rate     & 2.32E-03 & 2.36E-03 & 1.57\% \\
V3-WBTC & Liquidity Rate  & 5.75E-04 & 5.80E-04 & 0.85\% \\
\bottomrule
\end{tabular}
\end{table}

\begin{table}[H]
\centering
\caption{Interest Rate Volatility during Stress Periods}
\label{tab:volatility_stress} 
\renewcommand{\arraystretch}{1.2}
\begin{tabular}{p{2.5cm} p{2.3cm} p{3cm} p{2cm} p{2cm} p{1.5cm}}
\toprule
\textbf{Asset} & \textbf{Rate Type} & \textbf{Time Period} & \textbf{Aave Std.} & \textbf{TD3-BC Std.} & \textbf{Change} \\
\midrule
V2-WETH & Borrow Rate     & \textit{2022/2/1--2022/10/31} & 7.59E-02 & 7.64E-02 & +0.67\% \\
V2-WETH & Liquidity Rate  & \textit{2022/2/1--2022/10/31} & 4.07E-02 & 4.09E-02 & +0.38\% \\
V2-WETH & Borrow Rate     & \textit{2024/1/1--2024/12/31} & 1.18E-03 & 1.57E-03 & +32.69\% \\
V2-WETH & Liquidity Rate  & \textit{2024/1/1--2024/12/31} & 3.73E-04 & 5.91E-04 & +58.30\% \\
V3-WBTC & Borrow Rate     & \textit{2024/1/1--2024/12/31} & 2.49E-04 & 3.88E-04 & +56.11\% \\
V3-WBTC & Liquidity Rate  & \textit{2024/4/1--2024/10/31} & 5.44E-05 & 1.00E-04 & +84.49\% \\
\bottomrule
\end{tabular}
\end{table}